\pdfoutput=1

\documentclass[11pt]{article}

\usepackage[final]{acl}

\usepackage{times}
\usepackage{latexsym}

\usepackage[T1]{fontenc}

\usepackage[utf8]{inputenc}

\usepackage{microtype}

\usepackage{inconsolata}

\usepackage{graphicx}

\usepackage{subcaption}
\usepackage{graphicx}
\usepackage{amsmath,amssymb,amsthm}
\usepackage{booktabs}
\usepackage{dsfont}
\usepackage{multirow}
\usepackage{twemojis}
\usepackage{nicematrix}
\usepackage{enumitem}
\usepackage{mathtools}

\newcommand{\gold}{{\large\texttwemoji{1f642}}}
\newcommand{\lie}{{\large\texttwemoji{1f608}}}
\newcommand{\dstr}{{\large\texttwemoji{1f635-200d-1f4ab}}}

\newcommand{\orig}{{\large\texttwemoji{1f916}}}
\newcommand{\modi}{{\large\texttwemoji{1f527}}}

\definecolor{blue1}{HTML}{AEC7E8}
\definecolor{blue2}{HTML}{1F77B4}
\definecolor{orange1}{HTML}{FFBB78}

\newcommand{\cites}[1]{\citeauthor{#1}'s \citeyearpar{#1}}

\definecolor{darkgreen}{HTML}{005e19}
\definecolor{darkblue}{HTML}{240394}
\newcommand{\code}[1]{\texttt{#1}}
\newcommand{\exampleg}[1]{\textcolor{darkgreen}{\textbf{\small{\code{#1}}}}}
\newcommand{\examplegs}[1]{\textcolor{darkgreen}{\textbf{\scriptsize{\code{#1}}}}}
\newcommand{\exampleb}[1]{\textcolor{darkblue}{\textbf{\small{\code{#1}}}}}

\title{Llama See, Llama Do: A Mechanistic Perspective on Contextual Entrainment and Distraction in LLMs}

\author{
    Jingcheng Niu\thanks{Contribution during an internship at Microsoft Research.}$^{\dstr\lie}$, 
    Xingdi Yuan$^{\gold}$, 
    Tong Wang$^{\gold}$, 
    Hamidreza Saghir$^{\gold}$, 
    Amir H. Abdi$^{\gold}$\\
    $^{\dstr}$University of Toronto,\;
    $^{\lie}$UKP Lab, Technical University of Darmstadt,\;
    $^{\gold}$Microsoft\;\\
    \texttt{niu@cs.toronto.edu}, \; \texttt{\{eric.yuan,tong.wang,hsaghir,amirabdi\}@microsoft.com}
    }

\begin{document}
\maketitle
\begin{abstract}
    We observe a novel phenomenon, \emph{contextual entrainment}, across a wide range of language models (LMs) and prompt settings, providing a new mechanistic perspective on how LMs become distracted by ``irrelevant'' contextual information in the input prompt. Specifically, LMs assign significantly higher logits (or probabilities) to any tokens that have previously appeared in the context prompt, even for random tokens. This suggests that contextual entrainment is a \emph{mechanistic} phenomenon, occurring independently of the relevance or semantic relation of the tokens to the question or the rest of the sentence. We find statistically significant evidence that the magnitude of contextual entrainment is influenced by semantic factors. Counterfactual prompts have a greater effect compared to factual ones, suggesting that while contextual entrainment is a mechanistic phenomenon, it is modulated by semantic factors.
  
    We hypothesise that there is a circuit of attention heads --- the \emph{entrainment heads} --- that corresponds to the contextual entrainment phenomenon. Using a novel entrainment head discovery method based on differentiable masking, we identify these heads across various settings. When we ``turn off'' these heads, i.e., set their outputs to zero, the effect of contextual entrainment is significantly attenuated, causing the model to generate output that capitulates to what it would produce if no distracting context were provided.
    Our discovery of contextual entrainment, along with our investigation into LM distraction via the entrainment heads, marks a key step towards the mechanistic analysis and mitigation of the distraction problem.%
  \footnote{The code and data of this work can be found at \url{https://github.com/frankniujc/entrainment}.}
\end{abstract}

\section{Introduction}

Language models (LMs), especially large language models (LLMs), can sophisticatedly utilise contextual information provided in prompts to a surprising degree. \citet{brownLanguageModelsAre2020} was among the first to identify this capability and coin the term \emph{in-context learning} (ICL) to describe this capability. Subsequent work has demonstrated that LMs can process and utilise contextual information provided in prompts across various settings.

\begin{figure*}
  \centering
  \begin{subfigure}[t]{0.24\linewidth}
    \centering
    \includegraphics[width=\linewidth]{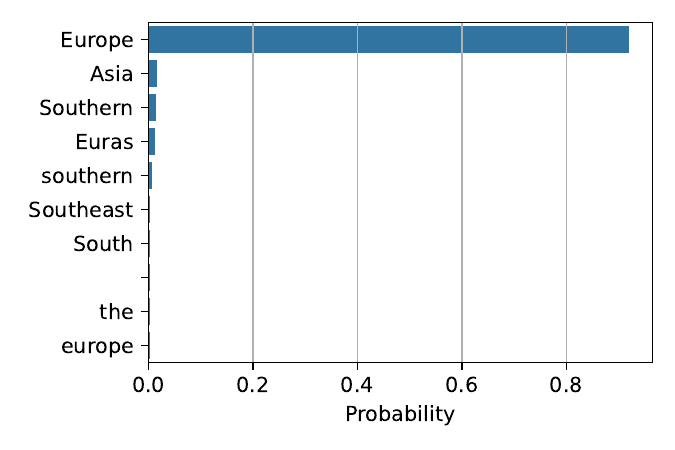}
    \parbox{0.93\linewidth}{
      \linespread{0.6}\selectfont
      \textcolor{darkgreen}{\textbf{\texttt{\scriptsize
        Greece is located on the continent of
      }}}%
    }
    \caption{When no context is provided to the model, it can confidently generate the correct response.}
    \label{fig:demo-no-context}
  \end{subfigure}
  \hfill
  \begin{subfigure}[t]{0.24\linewidth}
    \centering
    \includegraphics[width=\linewidth]{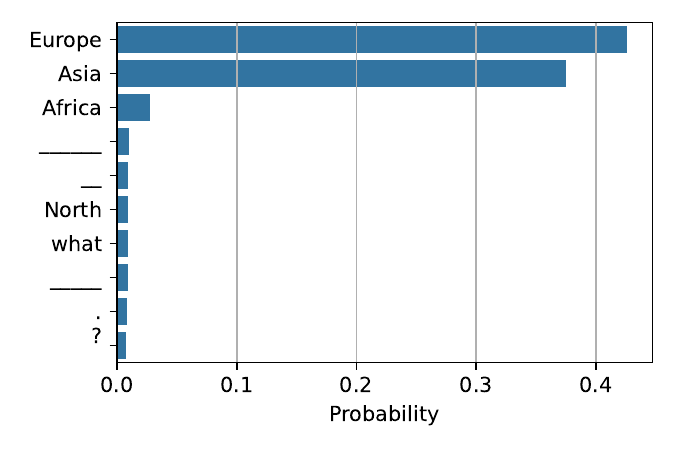}
    \parbox{0.93\linewidth}{
      \linespread{0.6}\selectfont
      \textcolor{darkgreen}{\textbf{\texttt{\scriptsize
      CONTEXT: Japan is in Asia. PROMPT: Greece is located on the continent of
      }}}%
    }
    \caption{However, when context from a related topic is provided, the model may assign a much higher probability to the distracting option.}
    \label{fig:demo-with-context}
  \end{subfigure}
  \hfill
  \begin{subfigure}[t]{0.24\linewidth}
    \centering
    \includegraphics[width=\linewidth]{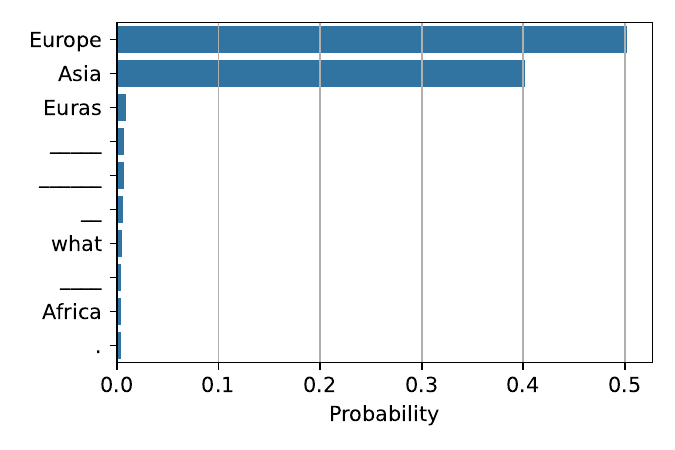}

    \parbox{0.93\linewidth}{
      \linespread{0.6}\selectfont
      \textcolor{darkgreen}{\textbf{\texttt{\scriptsize
        CONTEXT: Asia is the largest continent in the world by both land area and population. PROMPT: Greece is located on the continent of%
        }}}%
      }
    \caption{The model continues to exhibit distractions across different formats and paraphrasing of contextual information.}
    \label{fig:demo-reword-context}
  \end{subfigure}
  \hfill
  \begin{subfigure}[t]{0.24\linewidth}
    \centering
    \includegraphics[width=\linewidth]{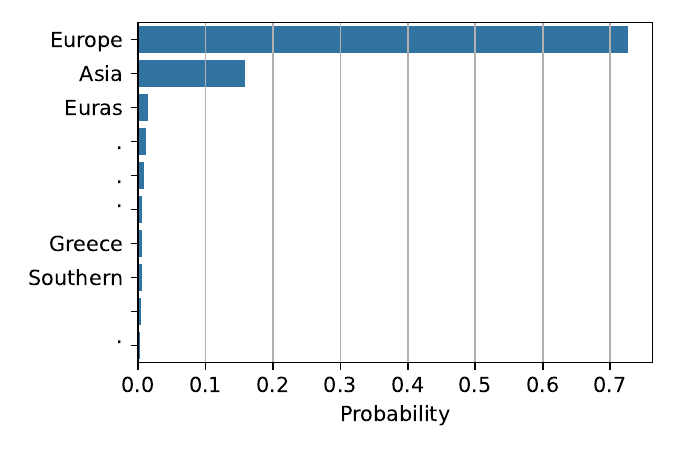}
    
    \parbox{0.93\linewidth}{
      \linespread{0.6}\selectfont
      \textcolor{darkgreen}{\textbf{\texttt{\scriptsize
      CONTEXT: Asia. PROMPT: Greece is located on the continent of
      }}}%
    }
    \caption{Just a single token, \emph{Asia}, can distract the model. This indicates that the phenomenon of LLM distraction is more superficial and fundamental than previously suggested.}
    \label{fig:demo-singletoken-context}
  \end{subfigure}
  \vspace{-0.5em}
  \caption{LLMs can be distracted by various types of context. Each sub-figure illustrates \exampleb{Llama-3.1-8B}'s output probability of the top ten tokens given the input prompt. The model inputs are displayed in \exampleg{green}.}
  \label{fig:demo}
  \vspace{-1em}
\end{figure*}

Nonetheless, LMs can also misuse contextual information in prompts  (Figure \ref{fig:demo}). \citet{shiLargeLanguageModels2023} experimented with inserting distracting, irrelevant information into grade-school maths problems and found that it successfully diverted the model from reaching the correct answer. Their work shed light on a fundamental issue within LMs, which they termed \emph{distraction}. Since then, distraction has been recognised as one of the most challenging and widespread issues for RAG \citep{yoranMakingRetrievalAugmentedLanguage2023,cuconasuPowerNoiseRedefining2024,wuHowEasilyIrrelevant2024}, prompting the development of distraction mitigation strategies. Notably, \citet{yoranMakingRetrievalAugmentedLanguage2023} proposed leveraging an NLI model to remove irrelevant context from the prompt as a solution to this problem.

Distraction, however, is a phenomenon in LMs that is easy to grasp but difficult to define precisely. Most prior work defines distraction using the term ``(ir)relevant,'' framed in RAG and information retrieval terms; i.e., whether the context prompt contains the information needed to answer the question correctly. While this provides an adequate general description of the problem, we identify challenges when examining it in greater detail. First, \emph{relevance} is too broad a concept. Consider the following context prompts: (1) \exampleg{Messi is a football player}, (2) \exampleg{Japan is in Asia}, (3) \exampleg{Greece is in Asia}, and (4) \exampleg{Colorless green ideas sleep furiously}. According to the earlier definition, all of these are ``irrelevant'' since they do not contain the information needed to correctly answer the question \exampleg{Greece is located on the continent of \rule{1em}{0.5pt}}. However, it is evident that they differ drastically in how they might influence an LM's response. Therefore, a more precise definition and a more fine-grained taxonomy of distraction are needed.
Moreover, we find evidence that these ``irrelevant'' context prompts can be beneficial in some cases, as they may provide useful implicit information about the question.

Regardless of the debate on the precise definition of distraction, we observe a phenomenon in LMs related to how they use and misuse information from the context prompt. In particular, we identify \emph{contextual entrainment}: LMs consistently assign significantly higher probabilities (or logits) to any tokens that have appeared earlier in the context prompt, regardless of their relevance or semantic relation to the question or the prompt. Simply put: \emph{llama see, llama do}. If a token appears in the context prompt, even a random one, the model assigns it a higher probability or logit. Our experiments show that various LMs exhibit contextual entrainment across a wide range of configurations.

At first glance, this phenomenon bears some resemblance to the inductive literal sequence copying phenomenon identified by \citet{elhageMathematicalFrameworkTransformer2021,olssonIncontextLearningInduction2022}, but the differences are substantial. First, contextual entrainment does not require the reappearance of a prefix as a trigger, unlike sequence copying, which depends on first encountering a pattern \exampleg{[A][B]} and then seeing the prefix \exampleg{[A]} again to generate \exampleg{[B]}. Instead, contextual entrainment occurs when a token has previously appeared in the context. Second, sequence copying is largely independent of semantic factors and token statistics \citep{olssonIncontextLearningInduction2022}, but the magnitude of contextual entrainment is influenced by semantic factors. In particular, we find that counterfactual prompts have a significantly stronger effect. We therefore identify contextual entrainment as a novel phenomenon that plays a crucial role in LMs' use (and misuse) of contextual information. It may be a major factor contributing to the distraction problem.

\begin{table*}
  \centering\small
  \setlength\tabcolsep{4pt}
  \begin{tabular}{l|p{4.2cm}|p{5.9cm}|cccccc} \toprule
    Context Setting & Context Prompt & Query Prompt & \multicolumn{3}{c|}{\dstr} & \multicolumn{3}{c}{\gold} \\
    \midrule
    Related & \examplegs{On the inside, bananas are white.} & \examplegs{What color are mangoes on the inside? They are} & \multicolumn{3}{c|}{\examplegs{white}} & \multicolumn{3}{c}{\examplegs{orange}} \\
    Irrelevant & \examplegs{The capital of Canada is Ottawa.} & \examplegs{What color are mangoes on the inside? They are} & \multicolumn{3}{c|}{\examplegs{Ottawa}} & \multicolumn{3}{c}{\examplegs{orange}} \\
    Random & \examplegs{Promotion} & \examplegs{What color are mangoes on the inside? They are} & \multicolumn{3}{c|}{\examplegs{Promotion}} & \multicolumn{3}{c}{\examplegs{orange}} \\
    \midrule
    \multicolumn{3}{c}{\it Distraction over Counterfactual Context} & \multicolumn{2}{|c|}{\lie} & \multicolumn{2}{c|}{\dstr} & \multicolumn{2}{c}{\gold} \\
    \midrule
    Counterfactual & \examplegs{On the inside, bananas are green.} & \examplegs{What color are mangoes on the inside? They are} & \multicolumn{2}{c|}{\examplegs{green}} & \multicolumn{2}{c|}{\examplegs{white}} & \multicolumn{2}{c}{\examplegs{orange}} \\
    \bottomrule
  \end{tabular}
  \caption{Prompt Setup. The emojis represent the target tokens: \lie: counterfactual; \dstr: distracting; \gold: correct.}
  \label{tab:prompt_settings}
  \vspace{-1em}
\end{table*}

Lastly, in contrast to the previous discussion on how this phenomenon differs from inductive literal sequence copying, here we identify a similarity. Similar to the induction heads identified by \citet{olssonIncontextLearningInduction2022}, we find that 3--10\% of the attention heads are associated with contextual entrainment, which we refer to as \emph{entrainment heads}. When these entrainment heads are disabled, i.e., their output is set to zero, the effect of contextual entrainment is drastically suppressed and the LM generates outputs similar to that produced when no context prompts are given. This provides an interesting insight into the mechanism governing the contextual entrainment phenomenon, and we hope our work can serve as a starting point for investigating the problem of distraction mechanistically.

\paragraph{Contributions} In this paper, we identify \textbf{contextual entrainment}, a novel phenomenon that plays a crucial role in LM distraction (\S\ref{sec:context_entrainment}). We provide evidence that the phenomenon can be considered mechanistic and occurs commonly across various LMs and settings (\S\ref{sub:experiment_results}). However, it is also influenced by semantic factors: counterfactual context more effectively induces contextual entrainment (\S\ref{sub:experiment_results}). Finally, we identify \textbf{entrainment heads} (\S\ref{sec:entrainment_heads}) using a novel method based on differentiable masking; when these heads are ``turned off,'' the effect of contextual entrainment is drastically suppressed.

\section{Related Work}

\paragraph{Distraction}
LMs are known to be susceptible to distractions caused by contextual information in prompts. For instance, \citet{shiLargeLanguageModels2023} found that while LMs can accurately solve grade-school maths problems, they may fail when provided with additional information in the prompt. They, however, did not explore the mechanisms underlying these distractions.
More research has since confirmed that LMs can be easily distracted \citep[\emph{inter alia}]{yoranMakingRetrievalAugmentedLanguage2023,wuHowEasilyIrrelevant2024,cuconasuPowerNoiseRedefining2024}. This problem has received remarkable attention in the RAG community due to its apparent connection to retrieval robustness. Since retrievers cannot always retrieve perfectly relevant documents, the LLM within a RAG system should be as resistant to distraction as possible.

\paragraph{Mechanistic Interpretability \& Induction Heads}

Our setup shares similarities with efforts to understand ICL, but, as discussed earlier, there are significant distinctions. \citet{elhageMathematicalFrameworkTransformer2021,olssonIncontextLearningInduction2022} successfully identified an inductive literal sequence copying phenomenon, a crucial step toward understanding how LMs perform ICL. They observed that if an LM has encountered a sequence of tokens --- even for random tokens --- it will repeat the sequence if it appears again in the prompt. For instance, given the input \exampleg{Category 40 ids node struction ... Category 40 ids node}, the model predicts \exampleg{struction} as the most probable next token. Therefore, they concluded that LMs possess some inductive capability and ``are not memorising a fixed table of n-gram statistics.'' They also identified certain attention heads as \emph{induction heads} in two-layer toy transformer models, which they claim perform pattern completion. More recently, \citet{crosbieInductionHeadsEssential2024} identified induction heads in real-world LMs such as \exampleb{Llama-3-8B}.

\section{Context Entrainment}
\label{sec:context_entrainment}

We present experiments that confirm the existence of the contextual entrainment phenomenon in this section. We find that contextual entrainment is a mechanistic phenomenon but is also influenced by semantic factors. The phenomenon is observed in all prompt settings, including those with completely randomly sampled tokens, suggesting that its existence is independent of semantic factors (i.e., contextual entrainment is mechanistic); however, the exact magnitude of its impact depends on semantic factors, as demonstrated by the significantly greater effect of counterfactual context.

\subsection{Experimental Setup}
\label{sub:experimental_setup}

\paragraph{Prompts \& Data}
Table~\ref{tab:prompt_settings} shows how we constructed our prompts using facts from the LRE dataset \citep{hernandezLinearityRelationDecoding2024}.
The LRE dataset contains facts in the triplet format: $\langle$\emph{source}, \emph{target}, \emph{relation}$\rangle$ or $\langle s$, $t$, $r \rangle$. For example, $\langle$\emph{Canada}, \emph{Ottawa}, \emph{capital}$\rangle$ corresponds to the fact that Canada's capital is Ottawa. Each fact in the dataset contains several prompt templates that we leverage to construct the context and query portion of the prompts.\footnote{Appendix \ref{app:lre_dataset} presents more details of the LRE dataset.}

\begin{figure*}[t!]

    \centering\small
    \begin{subfigure}[t]{\linewidth}
    \caption{A Review of the Experimental Setup. We provide the model with two types of input: (1) without context, and (2) with context. The context can take one of three forms (the counterfactual case will be discussed in the next section): \textbf{related}, \textbf{irrelevant}, or \textbf{random}. We observe the output (logit or probability) of the \textbf{correct} (\gold) and the \textbf{distracting} (\dstr) token.}
    \vspace{-1em}
    \begin{gather*}
        \overbrace{\exampleg{What is the capital of Canada? It is the city of}}^\text{Query Prompt}
        \exampleg{  }
        \begin{cases}
          \ell(\exampleg{Ottawa}) \text{: \gold{} without context} \\
          \ell(\exampleg{\textcolor{red}{Lima}}) \text{: \dstr{} without context}
        \end{cases}\\
        \overbrace{\exampleg{The capital of Peru is \textcolor{red}{Lima}.}}^\text{Context Prompt}
        \examplegs{  }
        \overbrace{\exampleg{What is the capital of Canada? It is the city of}}^\text{Query Prompt}
        \examplegs{  }
        \begin{cases}
          \ell(\exampleg{Ottawa}) \text{: \gold{} with context} \\
          \ell(\exampleg{\textcolor{red}{Lima}}) \text{: \dstr{} with context}
        \end{cases}
    \end{gather*}
    \end{subfigure}

    \begin{subfigure}[t]{0.32\linewidth}
        \scriptsize\centering
        \includegraphics[width=5cm]{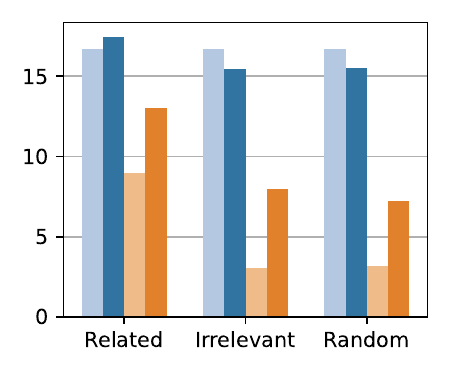}
        The bar height indicates the average logits of: \\
        \begin{tabular}{cc}
            \raisebox{0.7\height}{\colorbox{blue1}{\makebox(0.3ex,0.2ex){}}}: \gold{} without context & 
            \raisebox{0.7\height}{\colorbox{blue2}{\makebox(0.3ex,0.2ex){}}}: \gold{} with context
            \\
            \raisebox{0.7\height}{\colorbox{orange1}{\makebox(0.3ex,0.2ex){}}}: \dstr{} without context & 
            \raisebox{0.7\height}{\colorbox{orange}{\makebox(0.3ex,0.2ex){}}}: \dstr{} with context
        \end{tabular}
        \caption{Average logits for different tokens across prompt settings. The colour indicates the token type (blue: \gold{}, orange: \dstr), and the shade indicates the context setting (light: w/o context, dark: with context). \textbf{Observation}: The logits of the \gold{} tokens increase in the \textcolor{darkgreen}{\textbf{relevant}}$\uparrow$ setting and decrease in the \textcolor{red}{\textbf{irrelevant}}$\downarrow$ and \textcolor{red}{\textbf{random}}$\downarrow$ settings, while the logits of the \dstr{} tokens increase across \textcolor{darkgreen}{\textbf{all settings}}$\uparrow$.
        }
    \end{subfigure}\hfill
    \begin{subfigure}[t]{0.32\linewidth}
        \scriptsize\centering
        \includegraphics[width=5cm]{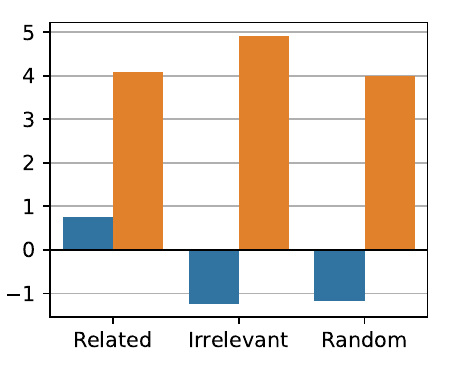}
        The bar height indicates the difference in \textbf{logits} with or without the distracting context across tokens:

        \raisebox{0.7\height}{\colorbox{blue2}{\makebox(0.3ex,0.2ex){}}}: $\Delta_{\ell}(\gold{}) = \ell(\gold{} \text{ w/ ctx.}) - \ell(\gold{} \text{ w/o ctx.}) $ 

        \raisebox{0.7\height}{\colorbox{orange}{\makebox(0.3ex,0.2ex){}}}: $\Delta_{\ell}(\dstr{}) = \ell(\dstr{} \text{ w/ ctx.}) - \ell(\dstr{} \text{ w/o ctx.}) $ 

        \caption{Now, let us zoom in on the magnitude of the increase or decrease: the difference in logits with or without the context across different tokens. Again, colour indicates token type (blue: \gold{}, orange: \dstr).
        \textbf{Observation}: The direction of change confirms the observation in (a). Notably, the magnitude of change for the \dstr{} tokens is greater than that of the \gold{} tokens.}
    \end{subfigure}\hfill
    \begin{subfigure}[t]{0.32\linewidth}
        \scriptsize\centering
        \includegraphics[width=5cm]{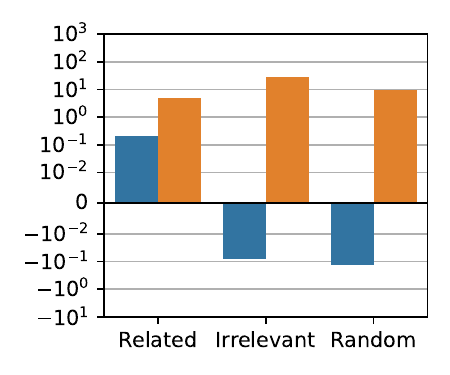}
        
        The bar height indicates the relative \textbf{probability} differences with or without the context across tokens:
        \begin{gather*}
        \text{\raisebox{0.7\height}{\colorbox{blue2}{\makebox(0.3ex,0.2ex){}}}:} \Delta_{p}(\gold{}) = \frac{p(\text{\gold{} w/ ctx}) - p(\text{\gold{} w/o ctx})}{p(\text{\gold{} w/o ctx})}\\
        \text{\raisebox{0.7\height}{\colorbox{orange}{\makebox(0.3ex,0.2ex){}}}:} \Delta_{p}(\dstr{}) = \frac{p(\text{\dstr{} w/ ctx}) - p(\text{\dstr{} w/o ctx})}{p(\text{\dstr{} w/o ctx})}
        \end{gather*}
        \vspace{-0.9em}
        \caption{The results observed in the logits carry over to the probability space. This subfigure shows the relative difference in probability. After applying softmax, we observe the same trend as in the logits.
        }
    \end{subfigure}

    \caption{Experimental Results for the \exampleb{Llama-3.1-8B} Model. Logit and probability values change consistently after the context prompt is provided to the model. The LM assigns significantly higher logits and probabilities to tokens that appear in the context prompts. All shifts are statistically significant, with $p<0.0001$ according to paired t-tests.}
    \label{fig:exp_distraction}
    \vspace{-1em}
\end{figure*}

We present the model with a context and a query in the prompt, separated by a single space character (e.g., \exampleg{<context> <query>}). Given a query generated from a fact $\langle s$, $t$, $r\rangle$, there are four context prompt settings: \textbf{related}, where facts $\langle s'$, $t'$, $r\rangle$ are sampled from the same relation type $r$ but differ in source $s'$ and target $t'$; \textbf{irrelevant}, where facts $\langle s'$, $t'$, $r'\rangle$ are sampled from a completely different relation type $r'$ without domain or range overlap; \textbf{random}, where the context consists of a single randomly chosen token; and \textbf{counterfactual}, where the target in the fact $\langle s'$, $t'_{\text{cf}}$, $r\rangle$ is replaced by another target sampled from the same relation. The random tokens are sampled from the Brown corpus \citep{francisBrownCorpusManual1979}. For larger relations that yield more than 100,000 combinations, we cap the size at 100,000 through random sampling.

Then, we feed in the input prompt (e.g., \exampleg{On the inside, bananas are \underline{white}. What color are mangoes on the inside? They are}), and the model outputs a logit score (or a probability after applying softmax) for every token in the vocabulary. We are interested in the logit or probability assigned to two (or three) tokens based on the input prompt: the \textbf{distracting} (\dstr) token (\exampleg{white}) that appears in the context prompt with a similar sentence format, and the \textbf{correct} (\gold) token (\exampleg{orange}) that answers the actual query prompt. In the counterfactual context prompt setting (e.g., \exampleg{On the inside, bananas are \underline{green}. What color are mangoes on the inside? They are}), we further distinguish between the \textbf{counterfactual} (\lie) token (\exampleg{green}) (which introduces incorrect information in the context prompt), the \textbf{distracting} (\dstr) token (\exampleg{white}) (which is the actual correct token in context), and the \textbf{correct} (\gold) token (\exampleg{orange}).

\paragraph{Language Models}
We experiment with \exampleb{GPT2 XL}~\citep{radfordLanguageModelsAre2019} and 4 LLaMA models~\citep{touvronLLaMAOpenEfficient2023}: \exampleb{Llama-3.1-8B}, \exampleb{Llama-3.1} \exampleb{-8B-Instruct}, \exampleb{Llama-2-7b-hf}, and \exampleb{Llama-2-13b-hf}.

\subsection{Experiment Results}
\label{sub:experiment_results}

\begin{figure}[!]
    \centering

    \begin{subfigure}{\linewidth}
        \includegraphics[width=2.5cm]{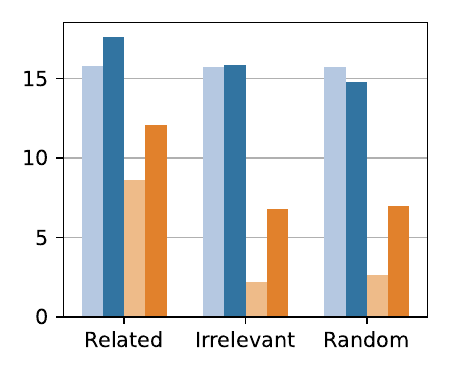}
        \includegraphics[width=2.5cm]{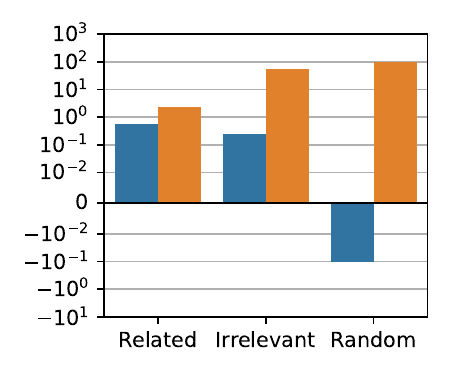}
        \includegraphics[width=2.5cm]{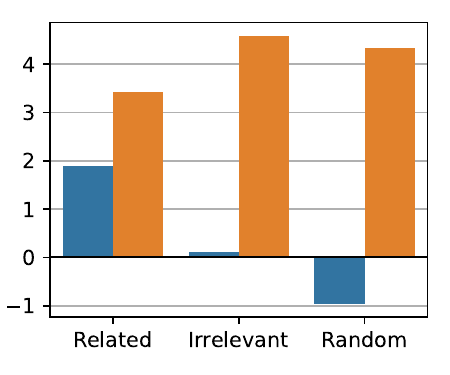}
        \caption{\exampleb{Llama-3.1-8B-Instruct}.}
    \end{subfigure}
    
    \begin{subfigure}{\linewidth}
        \includegraphics[width=2.5cm]{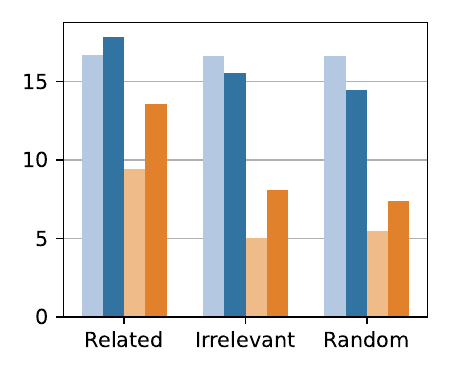}
        \includegraphics[width=2.5cm]{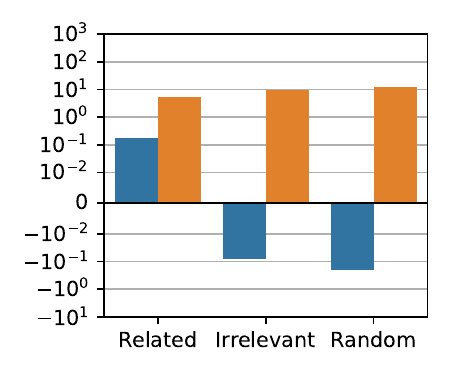}
        \includegraphics[width=2.5cm]{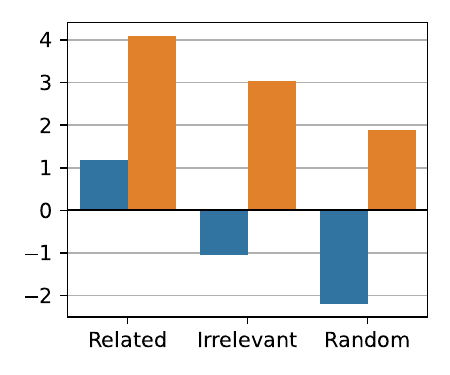} 
        \caption{\exampleb{Llama-2-7b-hf}.}
    \end{subfigure}
    
    \begin{subfigure}{\linewidth}
        \includegraphics[width=2.5cm]{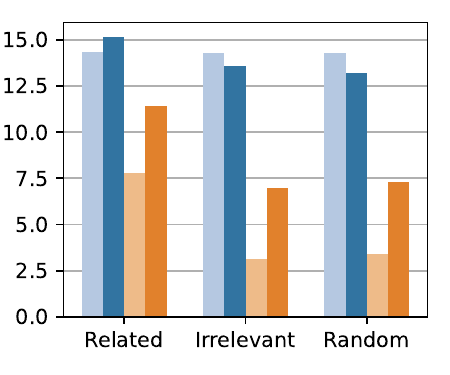}
        \includegraphics[width=2.5cm]{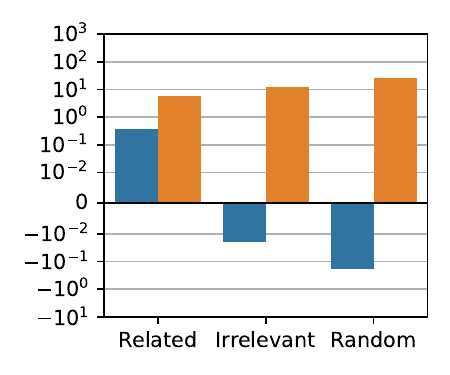}
        \includegraphics[width=2.5cm]{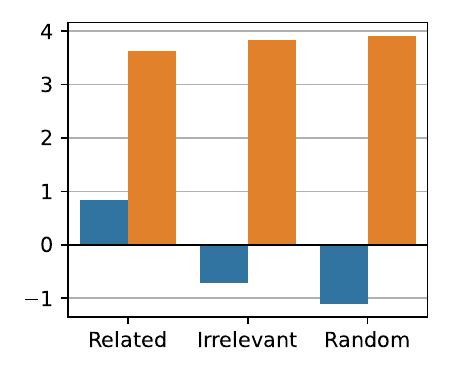}
        \caption{\exampleb{Llama-2-13b-hf}.}
    \end{subfigure}
    
    \begin{subfigure}{\linewidth}
        \includegraphics[width=2.5cm]{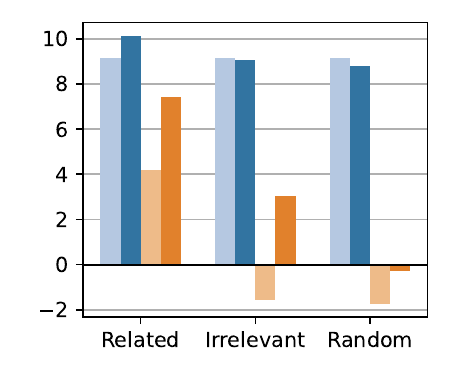}
        \includegraphics[width=2.5cm]{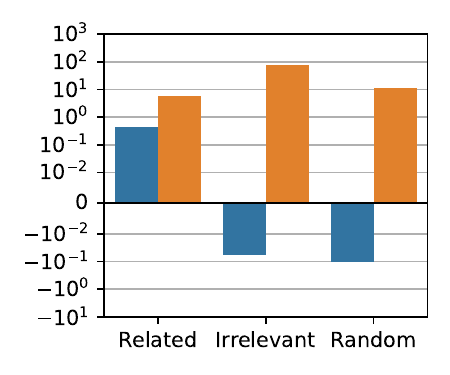}
        \includegraphics[width=2.5cm]{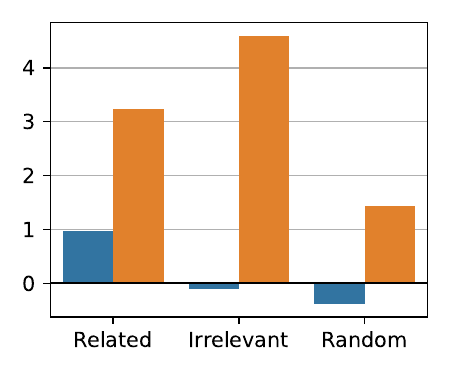}
        \caption{\exampleb{GPT-2 XL}.}
    \end{subfigure}

    \caption{Experimental Results for Other LMs. The figures follow the same plotting scheme as Figure~\ref{fig:exp_distraction}. The results confirm the same observations shown in Figure~\ref{fig:exp_distraction} and the findings discussed in Section~\ref{sub:experiment_results} across a wider range of models. Again, all shifts in probabilities and logits are statistically significant, with $p<0.0001$ according to paired t-tests.}
    \label{fig:results_more_models}
    \vspace{-1em}
\end{figure}

Figure \ref{fig:exp_distraction} presents the experimental results for three types of distracting contexts: distracting context from a related topic (Related), irrelevant topic (Irrelevant), and random token (Random), for the \exampleb{Llama-3.1-8B} model. The results of other LMs are shown in Figure~\ref{fig:results_more_models}, which support the same observations and findings. The figures show the averaged results across all LRE relations. The full list of experimental results, which supports the same conclusion, can be found in Appendix \ref{app:supp_exp_results}.

\paragraph{Finding 1: Contextual Entrainment: LMs assign higher logits and probabilities to tokens that appear in the context.}

When a model is given distracting context, there is a significant increase in the logits and probabilities of the corresponding distracting tokens. For example, when asked the question \exampleg{Greece is located in \rule{1em}{.4pt}}, distracting context prompts such as \exampleg{Japan is in Asia}, \exampleg{Bananas are yellow}, or even a single randomly sampled token, \exampleg{Promotion}, can cause the model to assign higher logits to the distracting tokens: \exampleg{Asia}, \exampleg{yellow}, and \exampleg{Promotion}, respectively. When normalised with softmax, logit increases translate into higher probabilities. Notably, the model typically assigns very low probabilities ($10^{-5}$ to $10^{-3}$) to these tokens, but with distracting context, their probabilities can increase by a factor of 10 to 100. Paired \cites{studentProbableErrorMean1908} \(t\)-tests confirm that these increases are statistically significant across all LMs, regardless of their size, family, or instruction-tuning status.

\paragraph{Finding 2: ``Distracting'' context prompts can be beneficial when relevant.}

While the probabilities and logits of the distracting tokens (\dstr) consistently increase, the direction of change for the correct token (\gold) varies based on topic relevance. Except for the instruction-tuned \exampleb{Llama-3.1-8B-Instruct} model,
there is a small but statistically significant decrease in the correct answer token's logit when context information from irrelevant or random context prompts is provided.

\begin{figure}
    \centering
    \includegraphics[width=\linewidth]{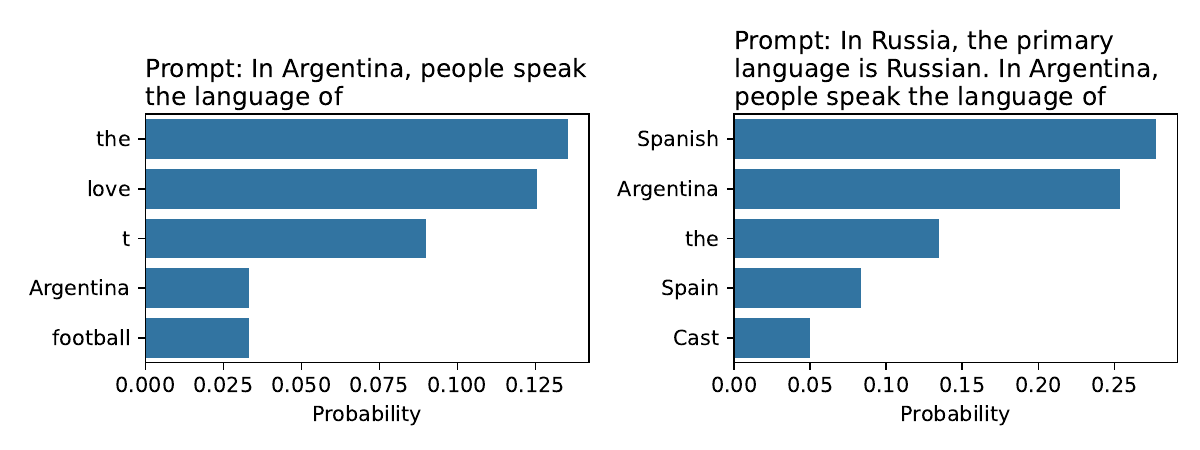}
    \vspace{-2em}
    \caption{Providing ``distracting'' yet relevant context can be beneficial. For instance, when a question is ambiguous, such context may help clarify its intended meaning or guide interpretation. For example, when ``distracting'' context is provided, \exampleb{Llama-3.1-8B}'s top responses to the prompt \exampleg{In Argentina, people speak the language of} shifted from \exampleg{love} and \exampleg{football} to the language-related tokens: \exampleg{Spanish}, \exampleg{Spain} and \exampleg{Cast} (the first wordpiece of ``Castilian Spanish'').}
    \label{fig:argentina_language}
    \vspace{-1em}
\end{figure}

While prior work typically groups ``irrelevant context'' into a single category and treats it as detrimental, our findings suggest the need for a more nuanced classification. Although distracting context may not contain the exact correct answer, the implicit hints it provides can be beneficial, increasing the likelihood that the model generates the correct response. Figure~\ref{fig:argentina_language} illustrates an example where ``distracting'' contextual information proves helpful, particularly in cases where the question is ambiguous. For instance, in the question \exampleg{In Argentina, people speak the language of} from the {\it country language} relation in the LRE dataset, the term \exampleg{language} can be interpreted metaphorically, leading to responses like \exampleg{love} or \exampleg{football}. However, the distracting context \exampleg{In Russia, the primary language is Russian} can help steer the model towards the correct answer.

\paragraph{Discussion: Contextual Entrainment --- A Novel Mechanistic Phenomenon}

Thus, a new perspective emerges for analysing the phenomenon of distraction from a mechanistic angle, which we term \emph{Contextual Entrainment}. Specifically, the model assigns a higher probability to tokens that appear within the context prompt. The fact that the model assigns higher probabilities to completely random tokens underscores the \emph{mechanistic} nature of this phenomenon, as no linguistic or factual factors can plausibly account for the increase in logits and probabilities of the random tokens.

Our coinage of the term \emph{Contextual Entrainment} does not imply any connection to human cognitive or psycholinguistic phenomena, such as brain entrainment \citep{poeppelSpeechRhythmsTheir2020,perezTimingBrainEntrainment2022} or lexical entrainment \citep{garrodSayingWhatYou1987,brennanConceptualPactsLexical1996}, nor does it suggest that LMs in any way replicate human brains or cognition. Rather, we use this term because \emph{entrainment} most accurately describes the phenomenon we have observed. It refers solely to the output patterns exhibited by LMs in response to contextual input, without making any claims about the underlying cognitive mechanisms or their resemblance to human cognitive processes.

\begin{figure}
\centering
\includegraphics[width=\linewidth]{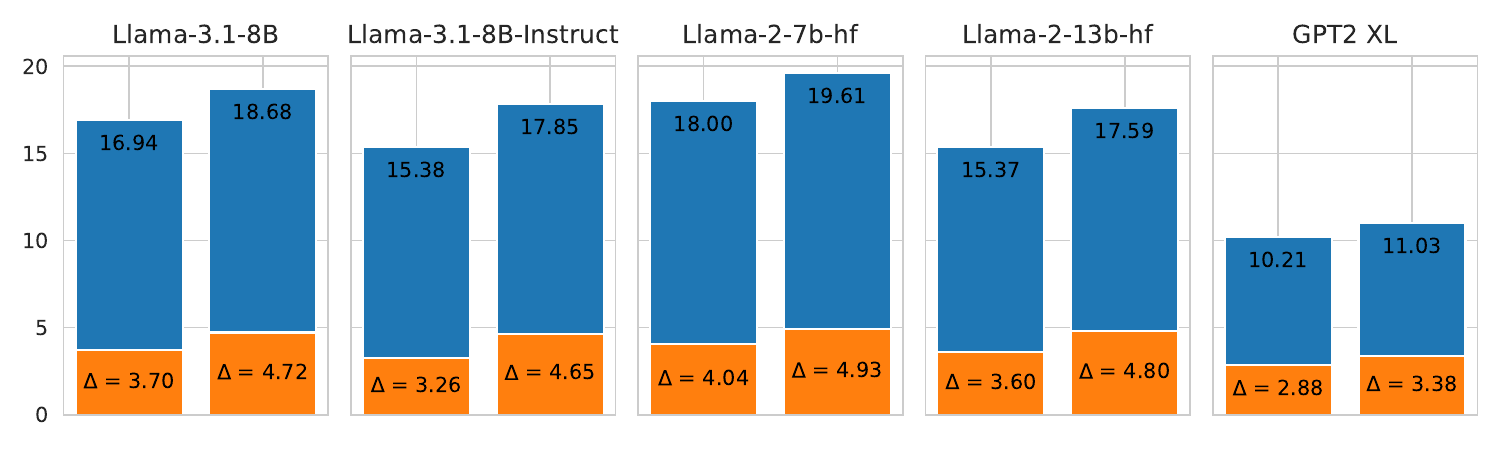}

  \setlength\tabcolsep{2.6pt}
  \begin{tabular}{p{0.22cm}p{.55cm}p{.55cm}p{.55cm}p{.55cm}p{.55cm}p{.55cm}p{.55cm}p{.55cm}p{.55cm}p{.55cm}}
    & \dstr & \lie & \dstr & \lie & \dstr & \lie & \dstr & \lie & \dstr & \lie 
  \end{tabular}
\vspace{-1em}
\caption{Counterfactual context prompts consistently cause greater distraction than factual context prompts.}
\label{fig:exp_lie}
  \vspace{-1em}
\end{figure}

\subsection{Counterfactual Experiment Results}
\label{sub:cf_experiment_results}

Figure \ref{fig:exp_lie} shows the results with counterfactual context prompts. Using these counterfactual context prompts results in a significantly stronger impact compared to previously identified factual context prompts. This suggests that, while we previously established that contextual entrainment is a ``mechanistic'' phenomenon, it is still subject to semantic factors in determining its magnitude of impact.

\paragraph{Finding 3: Counterfactual context prompts consistently cause stronger distraction than factual context prompts.}

We present the model with two types of distracting context prompts: a factual one (\exampleg{Japan is in Asia}) and a counterfactual one (\exampleg{Japan is in Africa}). Following the context prompt, we query the LM with a question (\exampleg{Greece is located in \rule{1em}{.4pt}}) and observe how it changes the logits and probabilities of the counterfactual \lie{} token (\exampleg{Africa}), the \dstr{} distracting token (\exampleg{Asia}), and the correct \gold{} token (\exampleg{Europe}).

The effect of distraction --- in other words, the magnitude of contextual entrainment --- is stronger for counterfactual context prompts than for factual prompts. This is evident because the absolute logits of the \dstr{} token when factual prompts are provided are significantly lower than those of the \lie{} token when counterfactual prompts are provided (height of blue bars in Figure~\ref{fig:exp_lie}). Moreover, the extent of change is greater for counterfactual prompts: with a factual prompt, the \dstr{} token's logits increase only slightly compared to no context, whereas a counterfactual prompt causes a much larger increase. This shows that counterfactual prompts create stronger distractions and greater shifts in the model's output (orange bar height in Figure~\ref{fig:exp_lie}).

\paragraph{Discussion: Contextual Entrainment --- A mechanistic phenomenon affected by semantic factors.}

We have established that the presence of contextual entrainment is independent of semantic factors, given that it occurs even with random tokens. However, in this subsection, we also find that this ``mechanistic'' phenomenon is nevertheless modulated by semantic factors. In particular, counterfactual prompts induce a greater effect on contextual entrainment than factual context prompts.

The mechanism through which LMs utilise information from prompts is not yet fully understood. There is an ongoing debate regarding whether this capability arises from mere memorisation \citep{golchinMemorizationInContextLearning2024} or from the implementation of an algorithm within the LMs weights and parameters during pre-training \citep{olssonIncontextLearningInduction2022,lindnerTracrCompiledTransformers2023}. Our findings suggest that this may not be a strict dichotomy; rather, it could be a compositional phenomenon in which both processes operate concurrently.

Furthermore, the fact that counterfactual prompts can cause stronger effects in contextual entrainment suggests that current models are particularly vulnerable to distraction from such inputs. This highlights the potential threat posed by dis- and misinformation.

\section{Entrainment Heads}
\label{sec:entrainment_heads}

Recent research presents the argument that attention heads play the crucial role in controlling the LMs' utilisation of context \citep[][\emph{inter alia}]{wangInterpretabilityWildCircuit2022,mengLocatingEditingFactual2022,jinCuttingHeadEnds2024,crosbieInductionHeadsEssential2024,yuMechanisticUnderstandingMitigation2024}.
Notably, \citet{jinCuttingHeadEnds2024} studied a similar phenomenon, termed \emph{knowledge conflicts}, which examines how models react when information from the context prompt contradicts the information acquired during pre-training. They use the terms \emph{internal memory} and \emph{external context} to refer to these two types of information. Furthermore, they identify two types of attention heads --- memory heads and context heads --- that correspond to the LM's utilisation of these distinct sources of information.

Knowledge conflict appears very similar to our counterfactual experiment; however, our research differs in several places. First and foremost, our prompt setting does not present a conflict. While we might both include a piece of counterfactual information in the context (e.g., \exampleg{The capital city of Germany is Moscow}), we will ask the model to answer a question unrelated to either Germany or Moscow (e.g., we would query the model with \exampleg{The capital of Nigeria is the city of \rule{1em}{0.5pt}}); whereas knowledge conflict would query the model with a question directly related to Germany or Moscow (e.g., \exampleg{The capital of Germany/Russia is the city of \rule{1em}{0.5pt}}). Second, counterfactual experiments are part of our investigation into the contextual entrainment phenomenon, whereas researchers who study information conflict focus solely on scenarios where such conflicts arise. Third, while identifying this novel phenomenon of contextual entrainment, we seek to understand the process by which LMs utilise information from the context prompt. In contrast, studies on information conflict focus more on the applicational aspect, where the desideratum is to find a way to ensure that LLMs can effectively resolve conflicting information and generate outputs that align with the intended factuality or coherence of the given context. Nevertheless, there are several aspects in which our research can mutually inform and benefit from one another.

In particular, \citet{jinCuttingHeadEnds2024} found that attention heads play a key role in utilising contextual information. As we will show later in this section, our experimental results further confirm this finding. However, their method is limited to investigating each attention head in isolation and does not consider the interaction between attention heads. Moreover, an increasing number of studies \citep{niuWhatDoesKnowledge2024,yuFunctionalFaithfulnessWild2024,bhaskarFindingTransformerCircuits2024} have identified issues with this individual approach, as it disregards the intricate structures of transformer LMs, and have advocated for a more holistic analysis of the entire computational ``circuit.'' Inspired by this line of research, we adapt the differentiable-masking-based approach proposed by \citet{yuFunctionalFaithfulnessWild2024,bhaskarFindingTransformerCircuits2024} to identify the set of attention heads responsible for contextual entrainment. Our approach yields better results than the method proposed by \citet{jinCuttingHeadEnds2024}, suggesting that a circuit of attention heads --- the ``entrainment heads'' --- may have been formed in the model to process contextual information.

\subsection{Entrainment Heads Discovery}

\begin{figure*}
  \centering
  \includegraphics[width=\linewidth]{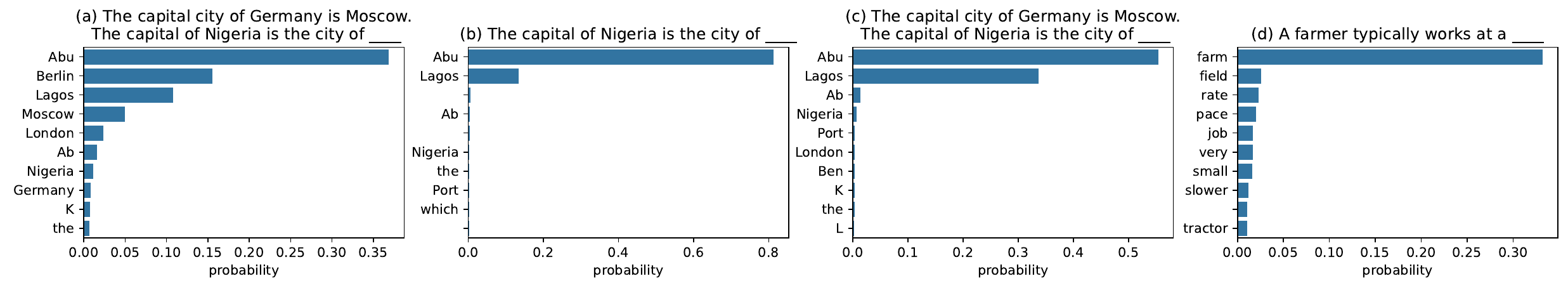}
  \vspace{-2em}
  $$
  \hspace{1.2em}
  \underbrace{\hspace{7.3cm}}_\text{Original Model \orig} \hspace{1em}
  \underbrace{\hspace{7.3cm}}_\text{Model with Entrainment Head Removed \modi}
  $$
  \vspace{-1em}
  \caption{Effects of ``Removing'' the Entrainment Heads. The figures show the top 10 token probabilities for their respective settings. (a,b) In the original model, when a piece of counterfactual information is presented, the model assigns higher probabilities to the distractions: \exampleg{Berlin} and \exampleg{Moscow}. (c) After setting the output of the identified entrainment heads to zero, however, the effect of contextual entrainment is drastically attenuated. (d)~This operation of removing the entrainment heads has only a small impact on other capabilities; the model can still correctly answer questions in other domains.}
  \label{fig:turn_off_heads}
  \vspace{-1em}
\end{figure*}

Inspired by \citet{yuFunctionalFaithfulnessWild2024}, we propose an automatic method to identify attention heads corresponding to contextual entrainment for each task configuration. We ``turn off'' specific heads by setting their contribution to the residual stream \citep{elhageMathematicalFrameworkTransformer2021}%
\footnote{We briefly review residual stream in Appendix \ref{app:residual_stream}.}
to zero. To achieve this, we introduce a binary mask $m_j$ for each head $h_j$, which selectively activates or deactivates heads ($\sum_{h_j\in H_i} m_j h_j (x_i)$).  
This mask is made differentiable by converting the sampled variable $s_i$ from a Gumbel-sigmoid distribution using the straight-through estimator \cite{bengioEstimatingPropagatingGradients2013}:
\begin{equation}\small
  s_i = \sigma\big(\frac{l_i - \log \frac{\log \mathcal{U}_1}{\log \mathcal{U}_2}}{\tau} \big); m_i = [\mathds{1}_{s_i > \frac{1}{2}} - s_i]_\text{\scriptsize detach} + s_i,
  \label{eq:eq4}
\end{equation}
where $ \tau \in (0, \infty) $ is a temperature hyperparameter, $ l_i $ is a learnable logit of the sigmoid distribution $ \sigma(\cdot) $, and $ \mathcal{U}_1, \mathcal{U}_2 \sim \text{Uniform}(0,1) $ are random variables drawn from a uniform distribution.

We can then apply gradient descent to a dataset to identify the optimal combinations of attention heads to disable in order to suppress contextual entrainment. Our objective is to determine the set of attention heads that contribute the most to contextual entrainment while minimising the number of heads used, using the following loss function:
\begin{equation}
    \small \mathcal{L} = \underbrace{\ell(\gold) - \ell(\dstr)}_\text{Logits $\Delta$} + \underbrace{\lambda \cdot \frac{1}{|H|}\sum_{i=1}^{|H|} \sigma(l_i)}_\text{Sparsity Loss} .
    \label{eq:loss}
\end{equation}

\paragraph{Experimental Setup}

We conduct our entrainment head experiments using \exampleb{Llama-3.1-8B}, which has 1,024 attention heads (32 layers × 32 heads). Each LRE relation is split into training (80\%), development (10\%), and test (10\%) sets. Entrainment heads are identified using the training set over 500 epochs,\footnote{We use the AdamW optimiser \citep{loshchilovDecoupledWeightDecay2019} with $\lambda = 1.0$, $\tau = 1.0$, and a learning rate of 1.0.} selecting the epoch with the best effect and fewest heads.\footnote{Specifically, we use the epoch with the maximum $\text{logit difference} + \text{number of heads}\times0.1$.} All results are reported on the test set, which the model has neither seen nor used for hyperparameter search or checkpoint selection.

\begin{table}
    \centering\scriptsize
    \begin{tabular}{c|cc|cc} \toprule
        \multirow{2}{*}{Measure} & \multicolumn{2}{c|}{\orig} & \multicolumn{2}{c}{\modi} \\
        & No \dstr & With \dstr & No \dstr & With \dstr \\ \midrule
        $\ell$(\gold) & 19.51 & 20.68 & 19.49 & 21.21 \\
        $\ell$(\dstr) & 8.75 & 12.99 & 7.87 & 8.01 \\
        $\Delta = \ell$(\gold) - $\ell$(\dstr) & 10.76 & 7.69 & 11.62 & 13.20 \\ \midrule
        Avg. \gold{} Token Rank & 1.00 & 1.00 & 1.00 & 1.00 \\
        Avg. \dstr{} Token Rank* & 1756.7 & 37.5 & 1707.3 & 1289.6 \\
        \bottomrule
    \end{tabular}
    \caption{Effects of ``Removing'' the Entrainment Heads across the Entire Country--Capital City Relation Test Set. Removing the entrainment heads caused a significant effect across logits delta and the ranks of the \dstr{} tokens, making them revert to the values observed when no distracting context is provided. *: $p<6.9\times10^{-54}$ according to paired t-tests between the \orig{} and \modi{} conditions.}
    \label{tab:country_capital_city_full}
  \vspace{-1em}
\end{table}

\subsection{Experiment Results \& Analysis}
\label{sub:case_study}

We first present our findings through the case study using the country--capital city relation. The remaining relations support the same findings; we will present them collectively at the end of this section.

\paragraph{``Turning off'' the entrainment heads drastically reduces contextual entrainment.}  
Our algorithm identified 36 entrainment heads for the country--capital city relation.%
\footnote{The entrainment heads identified are generally stable between runs, see Appendix~\ref{app:random}.}
When ``turning these entrainment heads off,'' i.e., setting their output to zero, the model attenuates the effect of contextual entrainment, as illustrated in Figure~\ref{fig:turn_off_heads}. Normally, \exampleb{Llama-3.1-8B} is distracted by the counterfactual context \exampleg{The capital city of Germany is Moscow}, confirming our previous findings in \S\ref{sub:cf_experiment_results}. However, after ``turning off'' the entrainment heads, the contextual entrainment effect is substantially attenuated. The rankings of the tokens \exampleg{Berlin} and \exampleg{Moscow} dropped from 2nd and 4th to 53rd and 68th, respectively. Additionally, the difference in logits between the correct \gold{} token (\exampleg{Abu})\footnote{The first wordpiece of \emph{Abuja}, the capital city of Nigeria.} and the distracting \dstr{} tokens (\exampleg{Berlin} and \exampleg{Moscow}) increased from 0.86 and 2.00 to 7.54 and 7.79, respectively. This attenuation is not a fluke. Table~\ref{tab:country_capital_city_full} shows that removing the entrainment heads significantly shifts the logits difference, probability difference, and the ranks of the distracting \dstr{} tokens towards the values observed when no distracting context is provided, across the entire country--capital city relation test set.

Table \ref{fig:entrainment_heads} shows that our differentiable-masking-based entrainment head discovery method applies to other relations in the LRE dataset, highlighting the generalisability of our approach. We observe an increase in logit differences on the same scale as our country--capital city case study, further supporting its findings. This result also suggests that entrainment heads are relation-specific.

\begin{table}
  \centering\small
  \setlength\tabcolsep{2pt}
    \begin{tabular}{lcccc} \toprule
    \multirow{2}{*}{Relation} & \multirow{2}{*}{\# Heads (Density)} & \multicolumn{3}{c}{$\ell$(\gold) - $\ell$(\dstr)} \\
    & & \orig & $\Rightarrow$ & \modi \\ \midrule
        company hq              & 90 (8.8\%)  & 3.94 & $\Rightarrow$ & 14.68 \\
        country capital city    & 36 (3.5\%)  & 7.69 & $\Rightarrow$ & 13.20 \\
        country currency        & 42 (4.1\%)  & 4.73 & $\Rightarrow$ & 11.67 \\
        country language        & 30 (2.9\%)  & 6.20 & $\Rightarrow$ & 8.95 \\
        country largest city    & 33 (3.2\%)  & 8.68 & $\Rightarrow$ & 13.35 \\
        food from country       & 38 (3.7\%)  & 3.98 & $\Rightarrow$ & 9.95 \\
        fruit inside color      & 56 (5.5\%)  & 0.97 & $\Rightarrow$ & 11.16 \\
        fruit outside color     & 80 (7.8\%)  & 2.14 & $\Rightarrow$ & 13.82 \\
        landmark in country     & 59 (5.8\%)  & 3.93 & $\Rightarrow$ & 9.68 \\
        landmark on continent   & 52 (5.1\%)  & 2.51 & $\Rightarrow$ & 9.14 \\
        product by company      & 110 (10.7\%) & 3.62 & $\Rightarrow$ & 16.47 \\
        star constellation name & 72 (7.0\%)  & 1.07 & $\Rightarrow$ & 8.87 \\
        task done by tool       & 66 (6.4\%)  & 4.70 & $\Rightarrow$ & 12.31 \\
        task person type        & 41 (4.0\%)  & 6.51 & $\Rightarrow$ & 12.47 \\
        work location           & 68 (6.6\%)  & 3.17 & $\Rightarrow$ & 12.68 \\
    \bottomrule
  \end{tabular}
  \caption{Entrainment Head Identified across All LRE Relations. A small set of attention heads (2.9\% to 10.7\%) can substantially increase the gap between the logits of \gold{} and \dstr{} tokens, i.e., attenuate contextual entrainment. This may suggest that these heads play a crucial role in the presence of contextual entrainment and can have broader implications for understanding how LMs utilise context information from prompts.}
  \label{fig:entrainment_heads}
  \vspace{-1em}
\end{table}

\paragraph{Entrainment heads are task-specific (or relation-specific), not model-specific.}

Again, as we have shown in Table~\ref{fig:entrainment_heads}, our method has identified a different set and a different number of entrainment heads for each LRE relation. This suggests that the entrainment heads are task-specific (or relation-specific), rather than model-specific. We do observe some level of overlap in the function of entrainment heads across different relations (Appendix~\ref{app:overlap_heads}).

This finding unveils a limitation of prior work that relies on individual approaches. Specifically, \citet{jinCuttingHeadEnds2024} classified attention heads as either memory heads or context heads, implying a strong and discrete assignment of functional roles to specific heads in a model-specific manner. This is an assumption that cannot account for our results.

\paragraph{Removing the entrainment heads has only a small effect on other LM capabilities.}  
While removing the entrainment heads significantly impacts contextual entrainment, it has only a negligible to small effect on other LM capabilities. First, we use other relations from the LRE dataset to evaluate whether the LM can still interpret the query and recall factual information, as well as perform ICL. Table~\ref{tab:no_other_effect} shows the performance of the original model (\orig) and the modified model (\modi) on all other relations without distracting context, demonstrating that the model can still perform factual recall.
We report both strict (the correct answer appears within the top-3 predicted tokens) and credulous (top-10) accuracy, as multiple correct answers may exist. Relying solely on whether the gold-standard token is the most probable leads to unstable results (Appendix~\ref{app:accuracy}). Moreover, we experiment with the three ICL tasks identified by \citet{brownLanguageModelsAre2020}: arithmetic, spelling correction, and translation (Appendix \ref{app:icl_tasks}). After removing the entrainment heads (\modi), the model exhibits only a small performance decrease (0.2$\sim$3\%) and continues to demonstrate strong ICL capabilities with high accuracy in the same ballpark as \orig. Appendix~\ref{app:more_effect} shows that removing the entrainment heads also has a small effect on the LMs' general capabilities.

\begin{table}
    \centering\scriptsize
    \begin{tabular}{l|cc|cc} \toprule
        \multirow{2}{*}{Relation} & \multicolumn{2}{c|}{Strict Acc.} & \multicolumn{2}{c}{Credulous Acc.} \\
         & \orig & \modi & \orig & \modi \\ \midrule
            company hq              & 83.5\%  & 90.0\%  & 88.0\%  & 90.0\%  \\
            country capital city    & 100.0\% & 100.0\% & 100.0\% & 100.0\% \\
            country currency        & 83.7\%  & 100.0\% & 100.0\% & 100.0\% \\
            country language        & 85.7\%  & 100.0\% & 100.0\% & 100.0\% \\
            country largest city    & 100.0\% & 100.0\% & 100.0\% & 100.0\% \\
            food from country       & 92.0\%  & 98.5\%  & 100.0\% & 98.5\%  \\
            fruit inside color      & 77.0\%  & 100.0\% & 98.0\%  & 100.0\% \\
            fruit outside color     & 38.0\%  & 84.0\%  & 82.0\%  & 84.0\%  \\
            landmark in country     & 89.5\%  & 91.0\%  & 95.0\%  & 91.0\%  \\
            landmark on continent   & 88.5\%  & 83.0\%  & 97.0\%  & 83.0\%  \\
            product by company      & 95.0\%  & 96.0\%  & 98.0\%  & 96.0\%  \\
            star constellation name & 84.7\%  & 89.3\%  & 92.3\%  & 89.3\%  \\
            task done by tool       & 78.0\%  & 91.0\%  & 93.5\%  & 91.0\%  \\
            task person type        & 78.5\%  & 80.0\%  & 80.0\%  & 80.0\%  \\
            work location           & 60.5\%  & 75.0\%  & 75.0\%  & 75.0\%  \\
            \midrule
            
            arithmetic 0-shot      & 100.0\% & 100.0\% & 100.0\% & 100.0\% \\
            spelling correction 1-shot       & 73.6\%  & 72.0\%  & 78.6\%  & 76.8\%  \\
            spelling correction 2-shot       & 94.6\%  & 91.6\%  & 97.0\%  & 94.8\%  \\
            spelling correction 5-shot       & 99.0\%  & 98.4\%  & 100.0\% & 100.0\% \\
            translation 1-shot      & 74.4\%  & 73.0\%  & 78.4\%  & 76.8\%  \\
            translation 2-shot      & 94.0\%  & 93.0\%  & 97.0\%  & 96.2\%  \\
            translation 5-shot      & 98.6\%  & 97.2\%  & 99.6\%  & 99.4\%  \\
        \bottomrule
    \end{tabular}
    \caption{Removing the entrainment heads of the country--capital city relation has a small to negligible effect on other LM capabilities. This table compares the strict (answer in top-3) and credulous (answer in top-10) accuracy of the original model (\orig) and the model with country--capital city entrainment heads removed (\modi). Removing these heads has a negligible effect on the LM's performance across other relations, with no obvious differences between \orig{} and \modi{}.}
    \label{tab:no_other_effect}
    \vspace{-1em}
\end{table}

This finding supports our hypothesis that this ``circuit'' of entrainment heads collectively corresponds to contextual entrainment rather than other capabilities and phenomena, such as factual recall, and is not strongly related to how LMs process and utilise contextual information or perform ICL more broadly. Thus, contextual entrainment and its connection to entrainment heads provide a novel perspective for understanding distraction. While current mitigation strategies \citep{yoranMakingRetrievalAugmentedLanguage2023,cuconasuPowerNoiseRedefining2024,wuHowEasilyIrrelevant2024} focus on methods external to the model --- either modifying the context prompt or prompting the model to self-correct through reasoning --- our findings suggest there could be a way to mitigate distraction by directly modifying or monitoring the internal mechanisms of LMs when performing RAG.

\section{Conclusion}

\emph{Llama see, llama do}. We observe and confirm a novel phenomenon, which we term \emph{contextual entrainment}. If a token has appeared previously in the prompt, the model assigns a higher logit to that token, even for random tokens. Thus, a novel mechanistic effect may be at play in governing how LMs process and utilise information from the prompt --- an effect that is analogous to but distinct from previously identified phenomena, such as the inductive literal sequence copying effect observed by \cites{olssonIncontextLearningInduction2022}. However, we also discover that contextual entrainment is influenced by semantic factors. This finding highlights the potential threat of dis- and misinformation, which may be more severe than mere mistakes generated by the model. It also suggests that there may not be a strict dichotomy between mechanistic and statistical interpretations of LMs.  
Our identification of the entrainment heads suggests that interpretability techniques could provide crucial insights for real-world applications, such as the study of distraction.


\section*{Limitations}

We did not conduct our experiments using larger LMs such as \exampleb{Llama-3.1-70B} and \exampleb{Llama-3.1-405B} due to resource limitations. However, we have used \exampleb{Llama-3.1-8B}, \exampleb{Llama-3.1-8B-Instruct}, and \exampleb{Llama-2-13B-hf}, as these models are sufficiently large and powerful for our experiments. Moreover, we observed no differences in findings between larger and smaller models, such as \exampleb{GPT-2 XL} and \exampleb{Llama-2-13B-hf}, suggesting that our results are not significantly affected by model scale within this range. We encourage others to reproduce our work using larger models to further validate our findings.

Our experimental setup is rigorous. However, since RAG is most relevant to the problem of distraction, we conducted experiments using only the LRE dataset in this setting. We did not use standard RAG datasets (e.g., SimpleQA \citep{weiMeasuringShortformFactuality2024}), as they are difficult to control and compare fairly. Nonetheless, our experiment with random token inputs provides strong evidence --- if such a setup yields successful results, then more structured approaches are unlikely to fail. Once again, we encourage others to reproduce our results using these datasets to further validate our findings.

Finally, while demonstrating two novel and insightful findings --- the contextual entrainment phenomenon and the identification of entrainment heads --- we do not propose an application to mitigate the distraction problem. We believe our findings serve as foundational steps toward addressing this issue. Given the depth of contributions presented in this work, we leave such applications for future research.

\section*{Acknowledgements}

We thank Alessandro Sordoni and Lucas Caccia (Microsoft Research) as well as Gerald Penn and Lei Yu (University of Toronto) for their valuable discussions and feedback.  We are especially grateful to Zhanao Fu (University of Toronto) for introducing us to relevant background on brain entrainment. We thank the reviewers and area chairs for their insightful suggestions and engaging discussions that helped improve this work. Finally, we thank the readers who reached out and pointed out the typos\footnote{The \href{https://aclanthology.org/2025.acl-long.791/}{ACL proceedings version} contains two typos, which we have corrected in this version. In Figure~\ref{fig:exp_distraction}, lighter colours indicate logits \emph{before} the distraction context is added, and darker colours indicate logits \emph{after} it is added. In Table~\ref{tab:country_capital_city_full}, the logit difference is defined as $\Delta = \ell$(\gold) - $\ell$(\dstr), not the other way around.} in the paper, which we have corrected in this version.


\bibliography{biblio}

\appendix

\newpage
\onecolumn

\begin{table*}[t!]
    \centering\small
    \begin{tabular}{lcp{5cm} p{5cm}} \toprule
        Relation & \# Samples & Context Templates & Query Templates \\ \midrule
        company hq & 674 & \exampleg{The headquarters of \{\} is in the city of}
        
        \exampleg{Where are the headquarters of \{\}? It is in the city of} & \exampleg{\{\} is headquartered in the city of}
        
        \exampleg{The headquarters of \{\} are in the city of} \\ \midrule 
        country capital city & 24 & \exampleg{The capital of \{\} is the city of}
        
        \exampleg{What is the capital of \{\}? It is the city of} & \exampleg{The capital city of \{\} is}
        
        \exampleg{The capital of \{\} is} \\ \midrule 
        country currency & 30 & \exampleg{What is the official currency of \{\}? It is called the}
        
        \exampleg{\{\}'s official currency is called the}
        
        \exampleg{The name of \{\}'s currency is the} & \exampleg{The official currency of \{\} is the}
        
        \exampleg{\{\}'s official currency is the} \\ \midrule 
        country language & 24 & \exampleg{\{\}, where most people speak}
        
        \exampleg{In \{\}, people speak the language of}
        
        \exampleg{People in \{\} speak the language of} & \exampleg{People in \{\} speak}
        
        \exampleg{The language used in \{\} is}
        
        \exampleg{In \{\}, the primary language is} \\ \midrule 
        country largest city & 24 & \exampleg{What is the largest city in \{\}? It is the city of}
        
        \exampleg{The largest city in \{\} is the city of} & \exampleg{The largest city in \{\} is}
        
        \exampleg{The biggest city in \{\} is} \\ \midrule 
        food from country & 30 & \exampleg{What is the country of origin for \{\}? It originates from}
        
        \exampleg{\{\} originates from the country of} & \exampleg{\{\} originates from}
        
        \exampleg{\{\} is from the country of} \\ \midrule 
        fruit inside color & 36 & \exampleg{What color are \{\} on the inside? They are} & \exampleg{On the inside, \{\} are} \\ \midrule 
        fruit outside color & 30 & \exampleg{What color are \{\} on the outside? They are the color of} & \exampleg{On the outside, \{\} are} \\ \midrule 
        landmark in country & 836 & \exampleg{What country is \{\} in? It is in}
        
        \exampleg{\{\} is in the country of} & \exampleg{\{\} is in the country of} \\ \midrule 
        landmark on continent & 947 & \exampleg{What continent is \{\} on? It is on}
        
        \exampleg{\{\} is on the continent of} & \exampleg{\{\} is on the continent of} \\ \midrule 
        product by company & 522 & \exampleg{Which company developed \{\}? It was developed by} & \exampleg{\{\} was created by}
        
        \exampleg{\{\} is a product of} \\ \midrule 
        star constellation name & 362 & \exampleg{What is the name of the constellation that \{\} is part of? It is part of}
        
        \exampleg{\{\} is part of the constellation named}
        
        \exampleg{What is the name of the constellation that \{\} belongs to? It belongs to} & \exampleg{\{\} is part of the constellation named} \\ \midrule 
        task person type & 32 & \exampleg{The task of \{\} would be best performed by someone with the role of a}
        
        \exampleg{The professional role most suited to handle \{\} is a} & \exampleg{\{\} is best suited for someone with the role of a} \\ \midrule 
        task done by tool & 52 & \exampleg{What tool is used for \{\}? Usually, you need a}
        
        \exampleg{To accomplish \{\}, you need a tool called a} & \exampleg{The tool used for \{\} is called a} \\ \midrule 
        work location & 38 & \exampleg{A \{\} typically works at a}
        
        \exampleg{You can usually find a \{\} working in a} & \exampleg{A \{\} typically works at a} \\ \midrule

    \end{tabular}
    \caption{Selected LRE Relations.}
    \label{tab:lre_rels}
\end{table*}

\begin{table*}[!]
    \centering\small
    \setlength\tabcolsep{3pt}
    \begin{tabular}{l|ccc|ccc|ccc|ccc} \toprule
    \multirow{2}{*}{Setting} & \multicolumn{3}{c|}{$\ell($\gold$)$} & \multicolumn{3}{c|}{$\ell($\dstr$)$} & \multicolumn{3}{c|}{$P($\gold$)$} & \multicolumn{3}{c}{$P($\dstr$)$} \\
    & \scriptsize No CTX & \scriptsize With CTX & $\Delta$ & \scriptsize No CTX & \scriptsize With CTX & $\Delta$ & \scriptsize No CTX & \scriptsize With CTX & $\Delta$ & \scriptsize No CTX & \scriptsize With CTX & $\Delta$ \\
      \midrule
      \multicolumn{13}{c}{\it Llama-3.1-8B } \\ 
      \midrule
    Distraction & 16.72 & 17.47 & 0.75 & 8.94 & 13.02 & 4.08 & 0.39 & 0.47 & 0.08 & 4.83e-03 & 0.03 & 0.02 \\
    Irrelevant & 16.68 & 15.45 & -1.23 & 3.05 & 7.96 & 4.92 & 0.38 & 0.35 & -0.03 & 8.44e-05 & 2.51e-03 & 2.42e-03 \\
    Random & 16.69 & 15.52 & -1.17 & 3.19 & 7.19 & 4.01 & 0.38 & 0.34 & -0.05 & 3.41e-04 & 3.53e-03 & 3.19e-03 \\
      \midrule
      \multicolumn{13}{c}{\it Llama-3.1-8B-Instruct } \\ 
      \midrule
    Distraction & 15.75 & 17.63 & 1.88 & 8.64 & 12.07 & 3.43 & 0.27 & 0.43 & 0.16 & 5.13e-03 & 0.02 & 0.01 \\
    Irrelevant & 15.74 & 15.84 & 0.11 & 2.19 & 6.78 & 4.58 & 0.27 & 0.34 & 0.07 & 1.31e-05 & 7.45e-04 & 7.32e-04 \\
    Random & 15.74 & 14.78 & -0.95 & 2.66 & 7.00 & 4.33 & 0.27 & 0.24 & -0.03 & 3.54e-05 & 3.55e-03 & 3.52e-03 \\
      \midrule
      \multicolumn{13}{c}{\it Llama-2-13b-hf } \\ 
      \midrule
    Distraction & 14.33 & 15.17 & 0.84 & 7.76 & 11.39 & 3.63 & 0.32 & 0.44 & 0.12 & 6.35e-03 & 0.04 & 0.04 \\
    Irrelevant & 14.28 & 13.56 & -0.72 & 3.14 & 6.98 & 3.84 & 0.32 & 0.31 & -6.20e-03 & 2.53e-04 & 3.35e-03 & 3.10e-03 \\
    Random & 14.29 & 13.19 & -1.10 & 3.41 & 7.32 & 3.91 & 0.32 & 0.26 & -0.06 & 3.58e-04 & 9.73e-03 & 9.37e-03 \\
      \midrule
      \multicolumn{13}{c}{\it Llama-2-7b-hf } \\ 
      \midrule
    Distraction & 16.67 & 17.87 & 1.19 & 9.44 & 13.54 & 4.10 & 0.40 & 0.47 & 0.07 & 5.67e-03 & 0.04 & 0.03 \\
    Irrelevant & 16.60 & 15.54 & -1.06 & 5.06 & 8.11 & 3.05 & 0.40 & 0.36 & -0.03 & 3.57e-04 & 3.74e-03 & 3.39e-03 \\
    Random & 16.61 & 14.43 & -2.18 & 5.50 & 7.39 & 1.89 & 0.40 & 0.32 & -0.08 & 4.27e-04 & 5.53e-03 & 5.10e-03 \\
      \midrule
      \multicolumn{13}{c}{\it GPT2 XL } \\ 
      \midrule
    Distraction & 9.17 & 10.13 & 0.96 & 4.18 & 7.41 & 3.23 & 0.18 & 0.26 & 0.08 & 7.76e-03 & 0.05 & 0.05 \\
    Irrelevant & 9.16 & 9.06 & -0.11 & -1.55 & 3.04 & 4.59 & 0.18 & 0.17 & -0.01 & 6.86e-05 & 5.28e-03 & 5.21e-03 \\
    Random & 9.16 & 8.79 & -0.37 & -1.72 & -0.28 & 1.44 & 0.18 & 0.16 & -0.02 & 1.84e-04 & 2.31e-03 & 2.13e-03 \\
    \bottomrule
    \end{tabular}
    \caption{Average Logits and Probabilities across All LRE Relations and Prompt Settings.}
    \label{tab:ce_all_avg}
\end{table*}

\begin{table*}
    \scriptsize\centering
    \setlength\tabcolsep{2pt}
    \begin{tabular}{l|l|ccc|ccc|ccc|ccc} \toprule
    \multirow{2}{*}{Relation}  & \multirow{2}{*}{Setting} & \multicolumn{3}{c|}{$\ell($\gold$)$} & \multicolumn{3}{c|}{$\ell($\dstr$)$} & \multicolumn{3}{c|}{$P($\gold$)$} & \multicolumn{3}{c}{$P($\dstr$)$} \\
    & & \scriptsize No CTX & \scriptsize With CTX & $\Delta$ & \scriptsize No CTX & \scriptsize With CTX & $\Delta$ & \scriptsize No CTX & \scriptsize With CTX & $\Delta$ & \scriptsize No CTX & \scriptsize With CTX & $\Delta$ \\ \midrule
 & dstr & 19.56 & 20.73 & 1.17 & 9.84 & 14.37 & 4.53 & 0.75 & 0.81 & 0.05 & 6.38e-03 & 0.03 & 0.02 \\
city in country & irr & 19.49 & 19.53 & 0.04 & 5.54 & 9.80 & 4.27 & 0.75 & 0.77 & 0.03 & 5.49e-05 & 2.38e-03 & 2.32e-03 \\
 & random & 19.52 & 16.96 & -2.56 & 3.68 & 7.47 & 3.79 & 0.75 & 0.63 & -0.12 & 2.33e-05 & 1.95e-03 & 1.93e-03 \\
 \midrule
 & dstr & 16.59 & 16.93 & 0.34 & 7.63 & 13.31 & 5.68 & 0.45 & 0.46 & 6.30e-03 & 1.50e-03 & 0.06 & 0.06 \\
company hq & irr & 16.66 & 15.43 & -1.23 & 4.34 & 9.64 & 5.30 & 0.46 & 0.36 & -0.09 & 1.22e-04 & 6.17e-03 & 6.05e-03 \\
 & random & 16.63 & 15.86 & -0.77 & 2.74 & 5.59 & 2.85 & 0.45 & 0.41 & -0.04 & 1.37e-04 & 3.26e-03 & 3.12e-03 \\
 \midrule
 & dstr & 18.42 & 20.03 & 1.60 & 7.48 & 11.95 & 4.47 & 0.83 & 0.91 & 0.08 & 1.47e-04 & 1.12e-03 & 9.73e-04 \\
country capital city & irr & 18.43 & 17.33 & -1.10 & 3.73 & 9.07 & 5.35 & 0.82 & 0.74 & -0.08 & 2.18e-05 & 1.42e-03 & 1.40e-03 \\
 & random & 18.44 & 17.29 & -1.16 & 3.59 & 8.68 & 5.09 & 0.82 & 0.74 & -0.09 & 3.28e-05 & 1.98e-03 & 1.95e-03 \\
 \midrule
 & dstr & 17.76 & 18.29 & 0.53 & 8.08 & 12.59 & 4.51 & 0.19 & 0.51 & 0.32 & 4.07e-04 & 7.58e-03 & 7.17e-03 \\
country currency & irr & 17.72 & 15.25 & -2.48 & 2.47 & 6.46 & 3.99 & 0.19 & 0.18 & -0.01 & 3.72e-04 & 2.00e-03 & 1.63e-03 \\
 & random & 17.73 & 16.41 & -1.33 & 3.30 & 6.53 & 3.24 & 0.19 & 0.18 & -8.41e-03 & 6.85e-05 & 1.20e-03 & 1.13e-03 \\
 \midrule
 & dstr & 14.79 & 18.32 & 3.53 & 8.29 & 12.14 & 3.85 & 0.23 & 0.56 & 0.34 & 8.41e-03 & 0.01 & 2.97e-03 \\
country language & irr & 14.79 & 17.50 & 2.71 & 3.04 & 8.10 & 5.05 & 0.23 & 0.56 & 0.34 & 8.68e-05 & 4.16e-03 & 4.07e-03 \\
 & random & 14.72 & 13.66 & -1.07 & 3.26 & 8.64 & 5.38 & 0.22 & 0.18 & -0.04 & 8.84e-04 & 0.01 & 0.01 \\
 \midrule
 & dstr & 18.11 & 19.68 & 1.58 & 7.38 & 11.16 & 3.78 & 0.67 & 0.79 & 0.12 & 1.22e-04 & 8.22e-04 & 7.00e-04 \\
country largest city & irr & 18.13 & 17.15 & -0.98 & 3.50 & 8.18 & 4.69 & 0.67 & 0.60 & -0.08 & 1.58e-05 & 8.34e-04 & 8.19e-04 \\
 & random & 18.13 & 17.53 & -0.60 & 3.73 & 9.40 & 5.67 & 0.67 & 0.64 & -0.04 & 1.29e-04 & 3.13e-03 & 3.00e-03 \\
 \midrule
 & dstr & 18.14 & 18.06 & -0.08 & 8.99 & 12.73 & 3.73 & 0.54 & 0.55 & 6.08e-03 & 1.70e-03 & 0.01 & 0.01 \\
food from country & irr & 18.12 & 16.25 & -1.87 & 4.03 & 9.89 & 5.86 & 0.54 & 0.41 & -0.13 & 3.28e-04 & 6.26e-03 & 5.93e-03 \\
 & random & 18.10 & 16.56 & -1.54 & 3.40 & 6.25 & 2.85 & 0.53 & 0.43 & -0.10 & 1.09e-04 & 6.43e-04 & 5.34e-04 \\
 \midrule
 & dstr & 15.78 & 15.57 & -0.21 & 13.35 & 14.58 & 1.23 & 0.12 & 0.17 & 0.05 & 0.02 & 0.07 & 0.05 \\
fruit inside color & irr & 15.77 & 12.82 & -2.95 & 1.54 & 6.28 & 4.74 & 0.12 & 0.10 & -0.02 & 7.85e-07 & 1.63e-04 & 1.62e-04 \\
 & random & 15.77 & 14.78 & -0.99 & 3.02 & 6.86 & 3.84 & 0.12 & 0.14 & 0.01 & 1.06e-03 & 1.31e-03 & 2.47e-04 \\
 \midrule
 & dstr & 14.57 & 14.53 & -0.03 & 11.47 & 12.05 & 0.58 & 0.06 & 0.08 & 0.02 & 9.13e-03 & 0.01 & 2.67e-03 \\
fruit outside color & irr & 14.51 & 12.97 & -1.55 & 3.40 & 8.52 & 5.12 & 0.06 & 0.05 & -5.39e-03 & 4.31e-06 & 2.00e-03 & 1.99e-03 \\
 & random & 14.52 & 12.95 & -1.57 & 3.58 & 7.92 & 4.35 & 0.06 & 0.04 & -0.02 & 1.77e-03 & 6.01e-03 & 4.24e-03 \\
 \midrule
 & dstr & 17.40 & 17.60 & 0.19 & 8.45 & 13.83 & 5.37 & 0.51 & 0.49 & -0.02 & 1.23e-03 & 0.04 & 0.04 \\
landmark in country & irr & 17.44 & 15.82 & -1.62 & 3.97 & 9.36 & 5.40 & 0.52 & 0.45 & -0.07 & 1.93e-04 & 6.79e-03 & 6.59e-03 \\
 & random & 17.42 & 15.79 & -1.62 & 3.30 & 6.55 & 3.25 & 0.51 & 0.43 & -0.08 & 1.61e-04 & 8.17e-04 & 6.56e-04 \\
 \midrule
 & dstr & 17.59 & 17.46 & -0.13 & 11.90 & 15.64 & 3.74 & 0.36 & 0.28 & -0.08 & 8.79e-03 & 0.07 & 0.06 \\
landmark on continent & irr & 17.06 & 15.59 & -1.47 & 4.15 & 8.50 & 4.36 & 0.32 & 0.24 & -0.09 & 2.39e-05 & 8.56e-04 & 8.32e-04 \\
 & random & 17.07 & 16.16 & -0.91 & 3.43 & 6.48 & 3.04 & 0.33 & 0.29 & -0.04 & 1.68e-04 & 5.98e-04 & 4.30e-04 \\
 \midrule
 & dstr & 15.99 & 15.03 & -0.96 & 6.83 & 11.13 & 4.31 & 0.49 & 0.47 & -0.02 & 1.95e-03 & 0.03 & 0.03 \\
product by company & irr & 16.03 & 14.29 & -1.74 & 2.67 & 5.43 & 2.77 & 0.50 & 0.41 & -0.08 & 7.22e-05 & 5.18e-04 & 4.45e-04 \\
 & random & 16.03 & 15.29 & -0.74 & 2.28 & 4.74 & 2.46 & 0.50 & 0.46 & -0.05 & 5.94e-04 & 1.30e-03 & 7.09e-04 \\
 \midrule
 & dstr & 17.57 & 16.33 & -1.24 & 11.83 & 15.43 & 3.60 & 0.28 & 0.26 & -0.02 & 4.51e-03 & 0.04 & 0.03 \\
star constellation name & irr & 17.56 & 14.94 & -2.62 & 2.88 & 7.87 & 4.98 & 0.28 & 0.20 & -0.07 & 2.74e-05 & 1.34e-03 & 1.31e-03 \\
 & random & 17.58 & 16.48 & -1.10 & 3.40 & 7.27 & 3.87 & 0.28 & 0.21 & -0.07 & 1.91e-04 & 2.13e-03 & 1.93e-03 \\
 \midrule
 & dstr & 15.82 & 16.56 & 0.74 & 6.65 & 12.38 & 5.73 & 0.28 & 0.34 & 0.06 & 1.60e-03 & 0.03 & 0.03 \\
task done by tool & irr & 15.82 & 13.67 & -2.15 & 0.83 & 5.32 & 4.50 & 0.28 & 0.19 & -0.09 & 5.52e-06 & 5.57e-04 & 5.51e-04 \\
 & random & 15.85 & 14.72 & -1.13 & 2.90 & 7.51 & 4.60 & 0.28 & 0.24 & -0.04 & 4.21e-05 & 6.53e-03 & 6.49e-03 \\
 \midrule
 & dstr & 14.71 & 16.92 & 2.21 & 6.70 & 11.91 & 5.21 & 0.21 & 0.39 & 0.18 & 9.63e-04 & 0.01 & 0.01 \\
task person type & irr & 14.70 & 14.32 & -0.38 & 0.99 & 6.74 & 5.75 & 0.21 & 0.20 & -6.79e-03 & 8.89e-06 & 1.70e-03 & 1.69e-03 \\
 & random & 14.73 & 13.90 & -0.84 & 2.87 & 8.11 & 5.24 & 0.21 & 0.17 & -0.04 & 3.44e-05 & 7.84e-03 & 7.80e-03 \\
 \midrule
 & dstr & 14.77 & 17.51 & 2.74 & 8.18 & 13.08 & 4.90 & 0.22 & 0.39 & 0.18 & 2.85e-03 & 0.04 & 0.03 \\
work location & irr & 14.70 & 14.34 & -0.35 & 1.68 & 8.23 & 6.55 & 0.22 & 0.20 & -0.02 & 1.31e-05 & 2.96e-03 & 2.94e-03 \\
 & random & 14.74 & 14.01 & -0.73 & 2.53 & 7.09 & 4.56 & 0.22 & 0.18 & -0.04 & 4.47e-05 & 2.89e-03 & 2.85e-03 \\
 \bottomrule

    \end{tabular}
    \caption{Average Logits and Probabilities for each LRE Relation. LM: \exampleb{Llama-3.1-8B}.}
    \label{tab:llama_31}
\end{table*}

\newpage
\twocolumn

\section{LRE Dataset}
\label{app:lre_dataset}

We construct our experimental prompts using commonsense and factual data from the LRE dataset \citep{hernandezLinearityRelationDecoding2024}. This dataset comprises 47 relations with over 10,000 instances, spanning four categories: factual associations, commonsense knowledge, implicit biases, and linguistic knowledge. The dataset was created by filtering out ambiguous or noisy triples from {\sc ParaRel} \citep{elazarMeasuringImprovingConsistency2021}, ensuring that each relation is well-defined and distinct. Additionally, the dataset was expanded by incorporating examples from structured knowledge sources, increasing coverage across various relation types.

However, during our experimentation, we identify several issues with certain relations. Some relations are too obscure for all of our chosen models to complete, resulting in highly noisy logits and probabilities for the correct tokens in some models. For example, the \emph{superhero archnemesis} relation includes instances such as $\langle$\emph{Martian Manhunter}, \emph{Despero}, \emph{superhero archnemesis}$\rangle$, which may not be well-represented across models. Additionally, some relations in the linguistics and bias domains are not particularly relevant to the LM distraction setting. As a result, we select 15 relations, with their statistics listed in Table~\ref{tab:lre_rels}.


\section{Context Entrainment Experiment Supplementary Results}
\label{app:supp_exp_results}
Table~\ref{tab:ce_all_avg} shows the results across all relations to generate Figure~\ref{fig:exp_distraction}. There is a small amount of variance for the logits of the correct token ($\ell(\text{\gold})$ No CTX) because we cap the number of samples to 100,000 by random sampling, as described in Section \ref{sub:experimental_setup}.
Table~\ref{tab:llama_31} shows the breakdown for each LRE relation for the \exampleb{Llama-3.1-8B} model.

\section{Background: The Residual Stream}
\label{app:residual_stream}

Here, we review the concept of the \emph{residual stream} \citep{elhageMathematicalFrameworkTransformer2021}, which provides the relevant background for our entrainment head discovery method. The basic idea is to view the computation of transformer LMs as maintaining a communication channel between model components via a shared vector space: the residual stream. Instead of operating independently, attention heads and MLP modules in each layer continuously pass and modify information through residual connections. Starting with the word embedding $x_0$, each layer (residual block) modifies it to become $x_i$, and the final result, $x_{-1}$, is converted into the output probability distribution by the unembed module. For each residual block at layer $i$, the output of the previous layer is $x_{i-1}$. Let $H_i$ denote the set of all attention heads in the layer. The block's output, $x_i$, is then computed as described in Equation~(\ref{eq:residual_stream}):
\begin{equation}
\begin{split}
x_i^{\text{\tiny mid}} = x_{i-1} + \sum_{h\in H_i} h(x_{i-1}); \\
x_i = x_i^{\text{\tiny mid}} + \mathrm{MLP}(x_i^{\text{\tiny mid}}).
\label{eq:residual_stream}
\end{split}
\end{equation}


\begin{table}[!]
    \centering\small
    \setlength\tabcolsep{2pt}
    \begin{tabular}{c|ccccc|c}
    \toprule
        Overlaps & Run 1 & Run 2 & Run 3 & Run 4 & Run 5 & \# Heads \\
    \midrule
        Run 1 & 1.000 & 0.381 & 0.542 & 0.372 & 0.516 & 96 (9.38\%) \\
        Run 2 & 0.381 & 1.000 & 0.419 & 0.548 & 0.322 & 60 (5.86\%) \\
        Run 3 & 0.542 & 0.419 & 1.000 & 0.500 & 0.351 & 89 (8.69\%) \\
        Run 4 & 0.372 & 0.548 & 0.500 & 1.000 & 0.317 & 70 (6.84\%) \\
        Run 5 & 0.516 & 0.322 & 0.351 & 0.317 & 1.000 & 92 (8.98\%) \\
    \bottomrule
    \end{tabular}
    \caption{Jaccard overlap between the five runs.}
    \label{tab:jaccard}
\end{table}

\begin{table*}
    \centering\small
    \begin{tabular}{l|cc|cc|cc} \toprule
        \multirow{2}{*}{Relation} & \multicolumn{2}{c|}{Exact Acc. (Top-1)} & \multicolumn{2}{c|}{Strict Acc. (Top-3)} & \multicolumn{2}{c}{Credulous Acc. (Top-10)} \\
         & \orig & \modi & \orig & \modi & \orig & \modi \\ \midrule
            company hq                 & 71.0\% & 63.5\% & 83.5\%  & 90.0\%  & 88.0\%  & 90.0\%  \\
            country capital city       & 94.0\% & 94.0\% & 100.0\% & 100.0\% & 100.0\% & 100.0\% \\
            country currency           & 18.0\% & 19.7\% & 83.7\%  & 100.0\% & 100.0\% & 100.0\% \\
            country language           & 37.0\% & 60.3\% & 85.7\%  & 100.0\% & 100.0\% & 100.0\% \\
            country largest city       & 97.0\% & 97.0\% & 100.0\% & 100.0\% & 100.0\% & 100.0\% \\
            food from country          & 78.0\% & 79.0\% & 92.0\%  & 98.5\%  & 100.0\% & 98.5\%  \\
            fruit inside color         & 49.0\% & 50.0\% & 77.0\%  & 100.0\% & 98.0\%  & 100.0\% \\
            fruit outside color        & 0.0\%  & 0.0\%  & 38.0\%  & 84.0\%  & 82.0\%  & 84.0\%  \\
            landmark in country        & 72.0\% & 54.5\% & 89.5\%  & 91.0\%  & 95.0\%  & 91.0\%  \\
            landmark on continent      & 41.5\% & 38.5\% & 88.5\%  & 83.0\%  & 97.0\%  & 83.0\%  \\
            product by company         & 86.0\% & 81.0\% & 95.0\%  & 96.0\%  & 98.0\%  & 96.0\%  \\
            star constellation name    & 22.3\% & 26.0\% & 84.7\%  & 89.3\%  & 92.3\%  & 89.3\%  \\
            task done by tool          & 58.0\% & 58.5\% & 78.0\%  & 91.0\%  & 93.5\%  & 91.0\%  \\
            task person type           & 67.0\% & 53.5\% & 78.5\%  & 80.0\%  & 80.0\%  & 80.0\%  \\
            work location              & 49.0\% & 49.0\% & 60.5\%  & 75.0\%  & 75.0\%  & 75.0\%  \\
            \midrule
            
            arithmetic 0-shot          & 95.8\% & 97.9\% & 100.0\% & 100.0\% & 100.0\% & 100.0\% \\
            spelling correction 1-shot & 58.2\% & 54.8\% & 73.6\%  & 72.0\%  & 78.6\%  & 76.8\%  \\
            spelling correction 2-shot & 77.6\% & 74.0\% & 94.6\%  & 91.6\%  & 97.0\%  & 94.8\%  \\
            spelling correction 5-shot & 87.4\% & 86.2\% & 99.0\%  & 98.4\%  & 100.0\% & 100.0\% \\
            translation 1-shot         & 58.4\% & 57.0\% & 74.4\%  & 73.0\%  & 78.4\%  & 76.8\%  \\
            translation 2-shot         & 74.8\% & 73.8\% & 94.0\%  & 93.0\%  & 97.0\%  & 96.2\%  \\
            translation 5-shot         & 85.4\% & 84.8\% & 98.6\%  & 97.2\%  & 99.6\%  & 99.4\%  \\
        \bottomrule
    \end{tabular}
    \caption{Removing the entrainment heads of the country--capital city relation has a small to negligible effect on other LM capabilities with exact (\gold{} must be top-1 response) included. The exact accuracy metric is highly unstable and therefore lacks reference value.}
    \label{tab:no_other_effect_with_exact}
\end{table*}

\section{Stochasticity between Runs}
\label{app:random}

The entrainment head discovery method involves several stochastic processes that can introduce noise to the results. Therefore, for the country--capital city relation, we re-run the entrainment head discovery process for 5 runs and report their Jaccard overlap. I.e., how many heads are identified by both runs divided by how many heads are identified by at least one run ($\frac{A\cap B}{A\cup B}$). Table~\ref{tab:jaccard} shows the Jaccard overlap result.

Overall, we observe a level of overlap substantially higher than random chance. Given that the five runs each end with 5.86–9.38\% of the attention heads selected, if the selection were random, the expected pairwise Jaccard overlap between any two runs would fall in the range of approximately 3.0\% to 4.9\%---substantially lower. This suggests a high degree of stability.

It is also interesting that the overlap is not perfect. In other words, we reproduce the ``hydra effect'' observed by \citet{mcgrathHydraEffectEmergent2023} (who found that when certain circuits are ablated, alternative ``backup'' circuits can emerge to take over the function) and the ``backup'' circuit phenomenon identified by \citet{wangInterpretabilityWildCircuit2022}. While we identified a circuit of entrainment heads with a strong causal effect on controlling the expression of contextual entrainment, it is also possible that these heads are part of a larger circuit, or that there are alternative components with competing functionalities. These are all interesting directions to explore in future work.

\section{Evaluation Method}
\label{app:accuracy}

We report both strict (the correct answer appears within the top-3 predicted tokens) and credulous (top-10) accuracy, as multiple correct answers may exist. Relying solely on whether the gold-standard token is the most probable leads to unstable results. For example, the LRE dataset includes the example \exampleg{On the outside, apples are \rule{1em}{0.5pt}}, where \exampleg{red} is the only correct answer. In some cases, however, the LM selects \exampleg{green} as the most probable next token, which is also a valid response.

Table~\ref{tab:no_other_effect_with_exact} shows Table~\ref{tab:no_other_effect}'s results with the exact (\gold{} must be top-1 response) metrics included. The exact accuracy metric is highly unstable due to the aforementioned reasons and therefore lacks reference value.

\section{Functional Overlap of Entrainment Heads}
\label{app:overlap_heads}

Table~\ref{fig:entrainment_heads_2} shows the effect of removing the country-capital entrainment heads compared to removing the task's heads. We can observe that removing the entrainment heads for country--capital relation causes a similar effect in suppressing contextual entrainment. This suppressing effect is, unsurprisingly, less consistent and smaller compared to removing the actual relation's entrainment heads.

Therefore, while there is some degree of cross-relation overlap in entrainment heads, it is still a relation-, task-, or domain-specific effect. We hypothesise that entrainment heads are part of the general task-solving mechanism or ``circuit.'' Recent work has highlighted a high level of idiosyncrasy in how LMs solve tasks. For example, \citet{liUnveilingPitfallsKnowledge2024}, \citet{niuWhatDoesKnowledge2024}, and \citet{berglundReversalCurseLLMs2024} found that LMs process ``A is B'' differently from ``B is A.'' Our observation aligns with these findings.

\begin{table*}[!]
  \centering\small
    \begin{tabular}{l|ccc|ccc} \toprule
    \multirow{3}{*}{Relation} & \multicolumn{3}{c|}{Removing Task} & \multicolumn{3}{c}{Removing Country--Capital} \\
                              & \multicolumn{3}{c|}{Entrainment Heads} & \multicolumn{3}{c}{Entrainment Heads} \\
    & \orig & $\Rightarrow$ & \modi & \orig & $\Rightarrow$ & \modi \\ \midrule
        company hq              & 3.94 & $\Rightarrow$ & 14.68 & 3.94 & $\Rightarrow$ & 6.81 \\
        country capital city    & 7.69 & $\Rightarrow$ & 13.20 & \multicolumn{3}{c}{---} \\
        country currency        & 4.73 & $\Rightarrow$ & 11.67 & 4.73 & $\Rightarrow$ & 9.04 \\
        country language        & 6.20 & $\Rightarrow$ & 8.95  & 6.20 & $\Rightarrow$ & 8.28 \\
        country largest city    & 8.68 & $\Rightarrow$ & 13.35 & 8.68 & $\Rightarrow$ & 13.27 \\
        food from country       & 3.98 & $\Rightarrow$ & 9.95  & 3.98 & $\Rightarrow$ & 8.41 \\
        fruit inside color      & 0.97 & $\Rightarrow$ & 11.16 & 0.97 & $\Rightarrow$ & 4.59 \\
        fruit outside color     & 2.14 & $\Rightarrow$ & 13.82 & 2.14 & $\Rightarrow$ & 7.45 \\
        landmark in country     & 3.93 & $\Rightarrow$ & 9.68  & 3.93 & $\Rightarrow$ & 6.14 \\
        landmark on continent   & 2.51 & $\Rightarrow$ & 9.14  & 2.51 & $\Rightarrow$ & 8.14 \\
        product by company      & 3.62 & $\Rightarrow$ & 16.47 & 3.62 & $\Rightarrow$ & 7.42 \\
        star constellation name & 1.07 & $\Rightarrow$ & 8.87  & 1.07 & $\Rightarrow$ & 5.95 \\
        task done by tool       & 4.70 & $\Rightarrow$ & 12.31 & 4.70 & $\Rightarrow$ & 6.53 \\
        task person type        & 6.51 & $\Rightarrow$ & 12.47 & 6.51 & $\Rightarrow$ & 8.21 \\
        work location           & 3.17 & $\Rightarrow$ & 12.68 & 3.17 & $\Rightarrow$ & 8.28 \\

    \bottomrule
  \end{tabular}
  \caption{Removing the entrainment heads for country-capital relation, for example, causes a similar effect in suppressing contextual entrainment.}
  \label{fig:entrainment_heads_2}
\end{table*}

\section{ICL Tasks}
\label{app:icl_tasks}
\begin{figure*}[!]
    \centering
    \includegraphics[width=\linewidth]{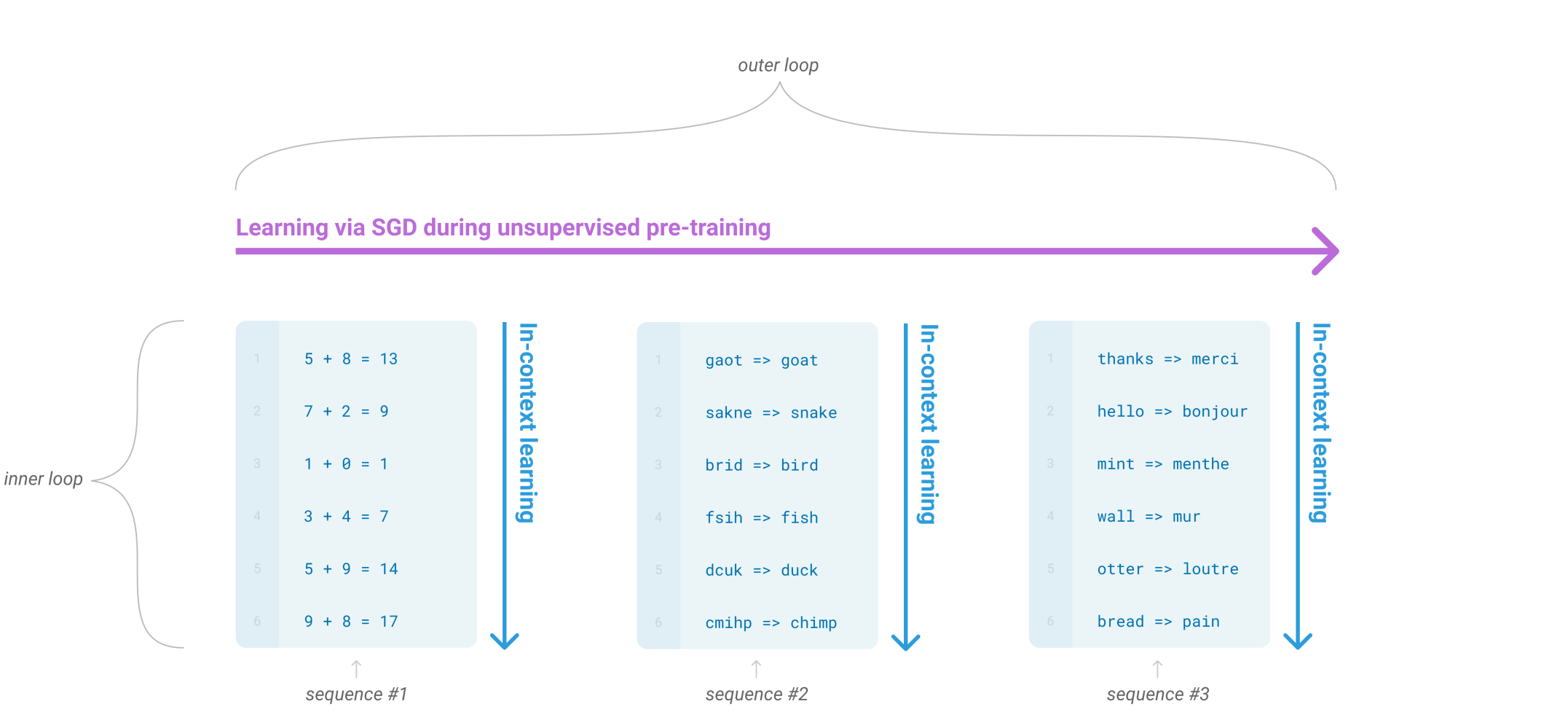}
    \caption{\cites{brownLanguageModelsAre2020} illustration of the three ICL tasks: arithmetic, spelling correction and translation.}
    \label{fig:icl_tasks}
\end{figure*}

We employ the three ICL tasks identified by \citet{brownLanguageModelsAre2020} to evaluate the effect of removing the entrainment heads on the LM's overall capability. In particular, \citet{brownLanguageModelsAre2020} proposed three tasks: arithmetic, spelling correction, and translation. Figure~\ref{fig:icl_tasks} shows \citet{brownLanguageModelsAre2020}'s illustration of these tasks. However, \citet{brownLanguageModelsAre2020} did not release the full dataset used for evaluation, so we recreate it.

\paragraph{Arithmetic} We randomly sampled 1000 prompts from all possible two digit summations. E.g., \exampleg{23 + 18 = 41}.

\newpage
\onecolumn
\begin{table*}
    \centering\small
    \begin{tabular}{p{14cm}}\toprule
    \multicolumn{1}{c}{\it (a) Prompt to Obtain the Spelling Correction Data} \\
    \midrule
    \exampleg{Give me 200 simple, random English words with 1 letter scrambled. For example:}

\exampleg{gaot => goat}

\exampleg{sakne => snake}

\exampleg{brid => bird}

\exampleg{fsih => fish}

\exampleg{dcuk => duck}

\exampleg{cmihp => chimp}

\exampleg{organize the results in a Python list:}

\exampleg{[('gaot', 'goat'), ('sakne', 'snake'), ...]} \\
\midrule

\multicolumn{1}{c}{\it (b) Prompt to Obtain the Translation Data} \\
\midrule
\exampleg{Give me 200 simple, random English words with their French translations. For example:}

\exampleg{thanks => merci}

\exampleg{hello => bonjour}

\exampleg{mint => menthe}

\exampleg{wall => mur}

\exampleg{otter => loutre}

\exampleg{bread => pain}

\exampleg{Avoid using French accent marks or letters that does exist in English}

\exampleg{organize the results in a Python list:
[('thanks', 'merci'), ('hello', 'bonjour'), ...]} \\
\bottomrule

    \end{tabular}
    \caption{Prompts Used to Obtain ICL Evaluation Data.}
    \label{tab:chatgpt_prompts}
\end{table*}

\begin{table*}
    \centering
    \begin{tabular}{p{15cm}}
         \toprule
         \exampleg{gaot} $\Rightarrow$ \exampleg{goat}, \exampleg{sakne} $\Rightarrow$ \exampleg{snake}, \exampleg{brid} $\Rightarrow$ \exampleg{bird}, \exampleg{fsih} $\Rightarrow$ \exampleg{fish}, \exampleg{dcuk} $\Rightarrow$ \exampleg{duck}, \exampleg{cmihp} $\Rightarrow$ \exampleg{chimp}, \exampleg{hosre} $\Rightarrow$ \exampleg{horse}, \exampleg{tiegr} $\Rightarrow$ \exampleg{tiger}, \exampleg{zbera} $\Rightarrow$ \exampleg{zebra}, \exampleg{muose} $\Rightarrow$ \exampleg{mouse}, \exampleg{lino} $\Rightarrow$ \exampleg{lion}, \exampleg{bera} $\Rightarrow$ \exampleg{bear}, \exampleg{wlof} $\Rightarrow$ \exampleg{wolf}, \exampleg{fxo} $\Rightarrow$ \exampleg{fox}, \exampleg{dree} $\Rightarrow$ \exampleg{deer}, \exampleg{frgo} $\Rightarrow$ \exampleg{frog}, \exampleg{girafef} $\Rightarrow$ \exampleg{giraffe}, \exampleg{doneky} $\Rightarrow$ \exampleg{donkey}, \exampleg{bunyn} $\Rightarrow$ \exampleg{bunny}, \exampleg{sehep} $\Rightarrow$ \exampleg{sheep}, \exampleg{whela} $\Rightarrow$ \exampleg{whale}, \exampleg{shrak} $\Rightarrow$ \exampleg{shark}, \exampleg{eagel} $\Rightarrow$ \exampleg{eagle}, \exampleg{corw} $\Rightarrow$ \exampleg{crow}, \exampleg{sawn} $\Rightarrow$ \exampleg{swan}, \exampleg{gosoe} $\Rightarrow$ \exampleg{goose}, \exampleg{pengiun} $\Rightarrow$ \exampleg{penguin}, \exampleg{ostirch} $\Rightarrow$ \exampleg{ostrich}, \exampleg{moneky} $\Rightarrow$ \exampleg{monkey}, \exampleg{kaola} $\Rightarrow$ \exampleg{koala}, \exampleg{haed} $\Rightarrow$ \exampleg{head}, \exampleg{hnad} $\Rightarrow$ \exampleg{hand}, \exampleg{foto} $\Rightarrow$ \exampleg{foot}, \exampleg{lge} $\Rightarrow$ \exampleg{leg}, \exampleg{amr} $\Rightarrow$ \exampleg{arm}, \exampleg{era} $\Rightarrow$ \exampleg{ear}, \exampleg{yee} $\Rightarrow$ \exampleg{eye}, \exampleg{lpi} $\Rightarrow$ \exampleg{lip}, \exampleg{teo} $\Rightarrow$ \exampleg{toe}, \exampleg{hiar} $\Rightarrow$ \exampleg{hair}, \exampleg{aplpe} $\Rightarrow$ \exampleg{apple}, \exampleg{baanna} $\Rightarrow$ \exampleg{banana}, \exampleg{ornage} $\Rightarrow$ \exampleg{orange}, \exampleg{maong} $\Rightarrow$ \exampleg{mango}, \exampleg{garpe} $\Rightarrow$ \exampleg{grape}, \exampleg{pecah} $\Rightarrow$ \exampleg{peach}, \exampleg{pera} $\Rightarrow$ \exampleg{pear}, \exampleg{plmu} $\Rightarrow$ \exampleg{plum}, \exampleg{kwii} $\Rightarrow$ \exampleg{kiwi}, \exampleg{leomn} $\Rightarrow$ \exampleg{lemon}, \exampleg{carrto} $\Rightarrow$ \exampleg{carrot}, \exampleg{pottao} $\Rightarrow$ \exampleg{potato}, \exampleg{onoin} $\Rightarrow$ \exampleg{onion}, \exampleg{tomtao} $\Rightarrow$ \exampleg{tomato}, \exampleg{ltetuce} $\Rightarrow$ \exampleg{lettuce}, \exampleg{rdaish} $\Rightarrow$ \exampleg{radish}, \exampleg{spianch} $\Rightarrow$ \exampleg{spinach}, \exampleg{cucubmer} $\Rightarrow$ \exampleg{cucumber}, \exampleg{ppeper} $\Rightarrow$ \exampleg{pepper}, \exampleg{celeyr} $\Rightarrow$ \exampleg{celery}, \exampleg{rde} $\Rightarrow$ \exampleg{red}, \exampleg{bleu} $\Rightarrow$ \exampleg{blue}, \exampleg{geren} $\Rightarrow$ \exampleg{green}, \exampleg{yellwo} $\Rightarrow$ \exampleg{yellow}, \exampleg{puprle} $\Rightarrow$ \exampleg{purple}, \exampleg{balck} $\Rightarrow$ \exampleg{black}, \exampleg{wihte} $\Rightarrow$ \exampleg{white}, \exampleg{pnik} $\Rightarrow$ \exampleg{pink}, \exampleg{borwn} $\Rightarrow$ \exampleg{brown}, \exampleg{gary} $\Rightarrow$ \exampleg{gray}, \exampleg{franec} $\Rightarrow$ \exampleg{france}, \exampleg{sapin} $\Rightarrow$ \exampleg{spain}, \exampleg{chnia} $\Rightarrow$ \exampleg{china}, \exampleg{inida} $\Rightarrow$ \exampleg{india}, \exampleg{itlay} $\Rightarrow$ \exampleg{italy}, \exampleg{jaapn} $\Rightarrow$ \exampleg{japan}, \exampleg{caanda} $\Rightarrow$ \exampleg{canada}, \exampleg{barzil} $\Rightarrow$ \exampleg{brazil}, \exampleg{geramny} $\Rightarrow$ \exampleg{germany}, \exampleg{russai} $\Rightarrow$ \exampleg{russia}, \exampleg{teaxs} $\Rightarrow$ \exampleg{texas}, \exampleg{manie} $\Rightarrow$ \exampleg{maine}, \exampleg{oiho} $\Rightarrow$ \exampleg{ohio}, \exampleg{iwoa} $\Rightarrow$ \exampleg{iowa}, \exampleg{uath} $\Rightarrow$ \exampleg{utah}, \exampleg{nevaad} $\Rightarrow$ \exampleg{nevada}, \exampleg{aalska} $\Rightarrow$ \exampleg{alaska}, \exampleg{haawii} $\Rightarrow$ \exampleg{hawaii}, \exampleg{flordia} $\Rightarrow$ \exampleg{florida}, \exampleg{gerogia} $\Rightarrow$ \exampleg{georgia}, \exampleg{tabel} $\Rightarrow$ \exampleg{table}, \exampleg{cahir} $\Rightarrow$ \exampleg{chair}, \exampleg{phoen} $\Rightarrow$ \exampleg{phone}, \exampleg{clcok} $\Rightarrow$ \exampleg{clock}, \exampleg{wacth} $\Rightarrow$ \exampleg{watch}, \exampleg{lihgt} $\Rightarrow$ \exampleg{light}, \exampleg{doro} $\Rightarrow$ \exampleg{door}, \exampleg{windwo} $\Rightarrow$ \exampleg{window}, \exampleg{sopon} $\Rightarrow$ \exampleg{spoon}, \exampleg{frok} $\Rightarrow$ \exampleg{fork}, \exampleg{rnu} $\Rightarrow$ \exampleg{run}, \exampleg{jmup} $\Rightarrow$ \exampleg{jump}, \exampleg{wlak} $\Rightarrow$ \exampleg{walk}, \exampleg{tlak} $\Rightarrow$ \exampleg{talk}, \exampleg{raed} $\Rightarrow$ \exampleg{read}, \exampleg{wrtie} $\Rightarrow$ \exampleg{write}, \exampleg{eta} $\Rightarrow$ \exampleg{eat}, \exampleg{slepe} $\Rightarrow$ \exampleg{sleep}, \exampleg{drvie} $\Rightarrow$ \exampleg{drive}, \exampleg{siwm} $\Rightarrow$ \exampleg{swim}, \exampleg{bgi} $\Rightarrow$ \exampleg{big}, \exampleg{smlal} $\Rightarrow$ \exampleg{small}, \exampleg{fsat} $\Rightarrow$ \exampleg{fast}, \exampleg{solw} $\Rightarrow$ \exampleg{slow}, \exampleg{hto} $\Rightarrow$ \exampleg{hot}, \exampleg{clod} $\Rightarrow$ \exampleg{cold}, \exampleg{nwe} $\Rightarrow$ \exampleg{new}, \exampleg{odl} $\Rightarrow$ \exampleg{old}, \exampleg{hpapy} $\Rightarrow$ \exampleg{happy}, \exampleg{sda} $\Rightarrow$ \exampleg{sad}, \exampleg{cta} $\Rightarrow$ \exampleg{cat}, \exampleg{dgo} $\Rightarrow$ \exampleg{dog}, \exampleg{cpu} $\Rightarrow$ \exampleg{cup}, \exampleg{pne} $\Rightarrow$ \exampleg{pen}, \exampleg{snu} $\Rightarrow$ \exampleg{sun}, \exampleg{mono} $\Rightarrow$ \exampleg{moon}, \exampleg{satr} $\Rightarrow$ \exampleg{star}, \exampleg{teer} $\Rightarrow$ \exampleg{tree}, \exampleg{rokc} $\Rightarrow$ \exampleg{rock}, \exampleg{blal} $\Rightarrow$ \exampleg{ball}, \exampleg{oepn} $\Rightarrow$ \exampleg{open}, \exampleg{clsoe} $\Rightarrow$ \exampleg{close}, \exampleg{puhs} $\Rightarrow$ \exampleg{push}, \exampleg{plul} $\Rightarrow$ \exampleg{pull}, \exampleg{lfit} $\Rightarrow$ \exampleg{lift}, \exampleg{dorp} $\Rightarrow$ \exampleg{drop}, \exampleg{crary} $\Rightarrow$ \exampleg{carry}, \exampleg{hodl} $\Rightarrow$ \exampleg{hold}, \exampleg{thorw} $\Rightarrow$ \exampleg{throw}, \exampleg{cacth} $\Rightarrow$ \exampleg{catch}, \exampleg{doctro} $\Rightarrow$ \exampleg{doctor}, \exampleg{laweyr} $\Rightarrow$ \exampleg{lawyer}, \exampleg{teacehr} $\Rightarrow$ \exampleg{teacher}, \exampleg{nusre} $\Rightarrow$ \exampleg{nurse}, \exampleg{drievr} $\Rightarrow$ \exampleg{driver}, \exampleg{artsit} $\Rightarrow$ \exampleg{artist}, \exampleg{sinegr} $\Rightarrow$ \exampleg{singer}, \exampleg{writre} $\Rightarrow$ \exampleg{writer}, \exampleg{chfe} $\Rightarrow$ \exampleg{chef}, \exampleg{pliot} $\Rightarrow$ \exampleg{pilot}, \exampleg{freind} $\Rightarrow$ \exampleg{friend}, \exampleg{enmey} $\Rightarrow$ \exampleg{enemy}, \exampleg{hosue} $\Rightarrow$ \exampleg{house}, \exampleg{hoem} $\Rightarrow$ \exampleg{home}, \exampleg{famliy} $\Rightarrow$ \exampleg{family}, \exampleg{moeny} $\Rightarrow$ \exampleg{money}, \exampleg{waetr} $\Rightarrow$ \exampleg{water}, \exampleg{frie} $\Rightarrow$ \exampleg{fire}, \exampleg{earht} $\Rightarrow$ \exampleg{earth}, \exampleg{widn} $\Rightarrow$ \exampleg{wind}, \exampleg{yse} $\Rightarrow$ \exampleg{yes}, \exampleg{on} $\Rightarrow$ \exampleg{no}, \exampleg{pu} $\Rightarrow$ \exampleg{up}, \exampleg{dwon} $\Rightarrow$ \exampleg{down}, \exampleg{lfet} $\Rightarrow$ \exampleg{left}, \exampleg{rigth} $\Rightarrow$ \exampleg{right}, \exampleg{ni} $\Rightarrow$ \exampleg{in}, \exampleg{otu} $\Rightarrow$ \exampleg{out}, \exampleg{dya} $\Rightarrow$ \exampleg{day}, \exampleg{ngiht} $\Rightarrow$ \exampleg{night}, \exampleg{ciyt} $\Rightarrow$ \exampleg{city}, \exampleg{tonw} $\Rightarrow$ \exampleg{town}, \exampleg{roda} $\Rightarrow$ \exampleg{road}, \exampleg{steret} $\Rightarrow$ \exampleg{street}, \exampleg{sohp} $\Rightarrow$ \exampleg{shop}, \exampleg{sotre} $\Rightarrow$ \exampleg{store}, \exampleg{bnak} $\Rightarrow$ \exampleg{bank}, \exampleg{csah} $\Rightarrow$ \exampleg{cash}, \exampleg{hosiptal} $\Rightarrow$ \exampleg{hospital}, \exampleg{clinci} $\Rightarrow$ \exampleg{clinic}, \exampleg{questoin} $\Rightarrow$ \exampleg{question}, \exampleg{anwser} $\Rightarrow$ \exampleg{answer}, \exampleg{probelm} $\Rightarrow$ \exampleg{problem}, \exampleg{solutoin} $\Rightarrow$ \exampleg{solution}, \exampleg{loev} $\Rightarrow$ \exampleg{love}, \exampleg{haet} $\Rightarrow$ \exampleg{hate}, \exampleg{peaec} $\Rightarrow$ \exampleg{peace}, \exampleg{wra} $\Rightarrow$ \exampleg{war}, \exampleg{truht} $\Rightarrow$ \exampleg{truth}, \exampleg{lei} $\Rightarrow$ \exampleg{lie}, \exampleg{muisc} $\Rightarrow$ \exampleg{music}, \exampleg{moive} $\Rightarrow$ \exampleg{movie}, \exampleg{boko} $\Rightarrow$ \exampleg{book}, \exampleg{paeg} $\Rightarrow$ \exampleg{page}, \exampleg{paepr} $\Rightarrow$ \exampleg{paper}, \exampleg{pecnil} $\Rightarrow$ \exampleg{pencil}, \exampleg{deks} $\Rightarrow$ \exampleg{desk}, \exampleg{soaf} $\Rightarrow$ \exampleg{sofa}, \exampleg{pillwo} $\Rightarrow$ \exampleg{pillow}, \exampleg{blnaket} $\Rightarrow$ \exampleg{blanket}. \\ 
         \bottomrule
    \end{tabular}
    \caption{200 Spelling Correction Pairs Used for ICL Capability Evaluation.}
    \label{tab:spell_corr}
\end{table*}

\begin{table*}
    \centering
    \begin{tabular}{p{15cm}}
         \toprule
    \exampleg{thanks} $\Rightarrow$ \exampleg{merci}, \exampleg{hello} $\Rightarrow$ \exampleg{bonjour}, \exampleg{mint} $\Rightarrow$ \exampleg{menthe}, \exampleg{wall} $\Rightarrow$ \exampleg{mur}, \exampleg{otter} $\Rightarrow$ \exampleg{loutre}, \exampleg{bread} $\Rightarrow$ \exampleg{pain}, \exampleg{water} $\Rightarrow$ \exampleg{eau}, \exampleg{friend} $\Rightarrow$ \exampleg{ami}, \exampleg{love} $\Rightarrow$ \exampleg{amour}, \exampleg{cat} $\Rightarrow$ \exampleg{chat}, \exampleg{dog} $\Rightarrow$ \exampleg{chien}, \exampleg{house} $\Rightarrow$ \exampleg{maison}, \exampleg{horse} $\Rightarrow$ \exampleg{cheval}, \exampleg{cow} $\Rightarrow$ \exampleg{vache}, \exampleg{cheese} $\Rightarrow$ \exampleg{fromage}, \exampleg{family} $\Rightarrow$ \exampleg{famille}, \exampleg{black} $\Rightarrow$ \exampleg{noir}, \exampleg{white} $\Rightarrow$ \exampleg{blanc}, \exampleg{red} $\Rightarrow$ \exampleg{rouge}, \exampleg{green} $\Rightarrow$ \exampleg{vert}, \exampleg{blue} $\Rightarrow$ \exampleg{bleu}, \exampleg{boy} $\Rightarrow$ \exampleg{garcon}, \exampleg{girl} $\Rightarrow$ \exampleg{fille}, \exampleg{night} $\Rightarrow$ \exampleg{nuit}, \exampleg{day} $\Rightarrow$ \exampleg{jour}, \exampleg{morning} $\Rightarrow$ \exampleg{matin}, \exampleg{evening} $\Rightarrow$ \exampleg{soir}, \exampleg{sun} $\Rightarrow$ \exampleg{soleil}, \exampleg{moon} $\Rightarrow$ \exampleg{lune}, \exampleg{star} $\Rightarrow$ \exampleg{etoile}, \exampleg{sky} $\Rightarrow$ \exampleg{ciel}, \exampleg{flower} $\Rightarrow$ \exampleg{fleur}, \exampleg{car} $\Rightarrow$ \exampleg{voiture}, \exampleg{city} $\Rightarrow$ \exampleg{ville}, \exampleg{country} $\Rightarrow$ \exampleg{pays}, \exampleg{beach} $\Rightarrow$ \exampleg{plage}, \exampleg{forest} $\Rightarrow$ \exampleg{foret}, \exampleg{river} $\Rightarrow$ \exampleg{riviere}, \exampleg{mountain} $\Rightarrow$ \exampleg{montagne}, \exampleg{desert} $\Rightarrow$ \exampleg{desert}, \exampleg{island} $\Rightarrow$ \exampleg{ile}, \exampleg{table} $\Rightarrow$ \exampleg{table}, \exampleg{chair} $\Rightarrow$ \exampleg{chaise}, \exampleg{window} $\Rightarrow$ \exampleg{fenetre}, \exampleg{door} $\Rightarrow$ \exampleg{porte}, \exampleg{book} $\Rightarrow$ \exampleg{livre}, \exampleg{pen} $\Rightarrow$ \exampleg{stylo}, \exampleg{pencil} $\Rightarrow$ \exampleg{crayon}, \exampleg{letter} $\Rightarrow$ \exampleg{lettre}, \exampleg{store} $\Rightarrow$ \exampleg{magasin}, \exampleg{restaurant} $\Rightarrow$ \exampleg{restaurant}, \exampleg{coffee} $\Rightarrow$ \exampleg{cafe}, \exampleg{tea} $\Rightarrow$ \exampleg{the}, \exampleg{juice} $\Rightarrow$ \exampleg{jus}, \exampleg{milk} $\Rightarrow$ \exampleg{lait}, \exampleg{egg} $\Rightarrow$ \exampleg{oeuf}, \exampleg{butter} $\Rightarrow$ \exampleg{beurre}, \exampleg{sugar} $\Rightarrow$ \exampleg{sucre}, \exampleg{salt} $\Rightarrow$ \exampleg{sel}, \exampleg{pepper} $\Rightarrow$ \exampleg{poivre}, \exampleg{chicken} $\Rightarrow$ \exampleg{poulet}, \exampleg{beef} $\Rightarrow$ \exampleg{boeuf}, \exampleg{fish} $\Rightarrow$ \exampleg{poisson}, \exampleg{bird} $\Rightarrow$ \exampleg{oiseau}, \exampleg{snake} $\Rightarrow$ \exampleg{serpent}, \exampleg{frog} $\Rightarrow$ \exampleg{grenouille}, \exampleg{turtle} $\Rightarrow$ \exampleg{tortue}, \exampleg{rabbit} $\Rightarrow$ \exampleg{lapin}, \exampleg{pig} $\Rightarrow$ \exampleg{cochon}, \exampleg{sheep} $\Rightarrow$ \exampleg{mouton}, \exampleg{goat} $\Rightarrow$ \exampleg{chevre}, \exampleg{fox} $\Rightarrow$ \exampleg{renard}, \exampleg{wolf} $\Rightarrow$ \exampleg{loup}, \exampleg{lion} $\Rightarrow$ \exampleg{lion}, \exampleg{tiger} $\Rightarrow$ \exampleg{tigre}, \exampleg{bear} $\Rightarrow$ \exampleg{ours}, \exampleg{phone} $\Rightarrow$ \exampleg{telephone}, \exampleg{computer} $\Rightarrow$ \exampleg{ordinateur}, \exampleg{keyboard} $\Rightarrow$ \exampleg{clavier}, \exampleg{screen} $\Rightarrow$ \exampleg{ecran}, \exampleg{mouse} $\Rightarrow$ \exampleg{souris}, \exampleg{camera} $\Rightarrow$ \exampleg{camera}, \exampleg{photo} $\Rightarrow$ \exampleg{photo}, \exampleg{movie} $\Rightarrow$ \exampleg{film}, \exampleg{music} $\Rightarrow$ \exampleg{musique}, \exampleg{song} $\Rightarrow$ \exampleg{chanson}, \exampleg{dance} $\Rightarrow$ \exampleg{danse}, \exampleg{poem} $\Rightarrow$ \exampleg{poeme}, \exampleg{library} $\Rightarrow$ \exampleg{bibliotheque}, \exampleg{museum} $\Rightarrow$ \exampleg{musee}, \exampleg{school} $\Rightarrow$ \exampleg{ecole}, \exampleg{university} $\Rightarrow$ \exampleg{universite}, \exampleg{teacher} $\Rightarrow$ \exampleg{professeur}, \exampleg{student} $\Rightarrow$ \exampleg{etudiant}, \exampleg{office} $\Rightarrow$ \exampleg{bureau}, \exampleg{job} $\Rightarrow$ \exampleg{travail}, \exampleg{money} $\Rightarrow$ \exampleg{argent}, \exampleg{bank} $\Rightarrow$ \exampleg{banque}, \exampleg{street} $\Rightarrow$ \exampleg{rue}, \exampleg{road} $\Rightarrow$ \exampleg{route}, \exampleg{building} $\Rightarrow$ \exampleg{batiment}, \exampleg{tall} $\Rightarrow$ \exampleg{grand}, \exampleg{small} $\Rightarrow$ \exampleg{petit}, \exampleg{short} $\Rightarrow$ \exampleg{court}, \exampleg{big} $\Rightarrow$ \exampleg{gros}, \exampleg{new} $\Rightarrow$ \exampleg{nouveau}, \exampleg{old} $\Rightarrow$ \exampleg{vieux}, \exampleg{happy} $\Rightarrow$ \exampleg{heureux}, \exampleg{sad} $\Rightarrow$ \exampleg{triste}, \exampleg{angry} $\Rightarrow$ \exampleg{fache}, \exampleg{tired} $\Rightarrow$ \exampleg{fatigue}, \exampleg{busy} $\Rightarrow$ \exampleg{occupe}, \exampleg{free} $\Rightarrow$ \exampleg{libre}, \exampleg{open} $\Rightarrow$ \exampleg{ouvert}, \exampleg{closed} $\Rightarrow$ \exampleg{ferme}, \exampleg{expensive} $\Rightarrow$ \exampleg{couteux}, \exampleg{cheap} $\Rightarrow$ \exampleg, \exampleg{yes} $\Rightarrow$ \exampleg{oui}, \exampleg{no} $\Rightarrow$ \exampleg{non}, \exampleg{maybe} $\Rightarrow$ \exampleg,  \exampleg{never} $\Rightarrow$ \exampleg{jamais}, \exampleg{always} $\Rightarrow$ \exampleg{toujours}, \exampleg{often} $\Rightarrow$ \exampleg{souvent}, \exampleg{sometimes} $\Rightarrow$ \exampleg{parfois}, \exampleg{rarely} $\Rightarrow$ \exampleg{rarement}, \exampleg{early} $\Rightarrow$ \exampleg{tot}, \exampleg{late} $\Rightarrow$ \exampleg{tard}, \exampleg{now} $\Rightarrow$ \exampleg{maintenant}, \exampleg{soon} $\Rightarrow$ \exampleg{bientot}, \exampleg{yesterday} $\Rightarrow$ \exampleg{hier}, \exampleg{today} $\Rightarrow$ \exampleg{aujourd}), \exampleg{tomorrow} $\Rightarrow$ \exampleg{demain}, \exampleg{hour} $\Rightarrow$ \exampleg{heure}, \exampleg{minute} $\Rightarrow$ \exampleg{minute}, \exampleg{second} $\Rightarrow$ \exampleg{seconde}, \exampleg{time} $\Rightarrow$ \exampleg{temps}, \exampleg{moment} $\Rightarrow$ \exampleg{moment}, \exampleg{week} $\Rightarrow$ \exampleg{semaine}, \exampleg{month} $\Rightarrow$ \exampleg{mois}, \exampleg{year} $\Rightarrow$ \exampleg{annee}, \exampleg{monday} $\Rightarrow$ \exampleg{lundi}, \exampleg{tuesday} $\Rightarrow$ \exampleg{mardi}, \exampleg{wednesday} $\Rightarrow$ \exampleg{mercredi}, \exampleg{thursday} $\Rightarrow$ \exampleg{jeudi}, \exampleg{friday} $\Rightarrow$ \exampleg{vendredi}, \exampleg{saturday} $\Rightarrow$ \exampleg{samedi}, \exampleg{sunday} $\Rightarrow$ \exampleg{dimanche}, \exampleg{spring} $\Rightarrow$ \exampleg{printemps}, \exampleg{summer} $\Rightarrow$ \exampleg{ete}, \exampleg{autumn} $\Rightarrow$ \exampleg{automne}, \exampleg{winter} $\Rightarrow$ \exampleg{hiver}, \exampleg{police} $\Rightarrow$ \exampleg{police}, \exampleg{fire} $\Rightarrow$ \exampleg{feu}, \exampleg{help} $\Rightarrow$ \exampleg{aide}, \exampleg{problem} $\Rightarrow$ \exampleg{probleme}, \exampleg{question} $\Rightarrow$ \exampleg{question}, \exampleg{answer} $\Rightarrow$ \exampleg{reponse}, \exampleg{truth} $\Rightarrow$ \exampleg{verite}, \exampleg{lie} $\Rightarrow$ \exampleg{mensonge}, \exampleg{idea} $\Rightarrow$ \exampleg{idee}, \exampleg{important} $\Rightarrow$ \exampleg{important}, \exampleg{interesting} $\Rightarrow$ \exampleg{interessant}, \exampleg{possible} $\Rightarrow$ \exampleg{possible}, \exampleg{impossible} $\Rightarrow$ \exampleg{impossible}, \exampleg{difficult} $\Rightarrow$ \exampleg{difficile}, \exampleg{easy} $\Rightarrow$ \exampleg{facile}, \exampleg{strong} $\Rightarrow$ \exampleg{fort}, \exampleg{weak} $\Rightarrow$ \exampleg{faible}, \exampleg{light} $\Rightarrow$ \exampleg{lumiere}, \exampleg{dark} $\Rightarrow$ \exampleg{sombre}, \exampleg{direction} $\Rightarrow$ \exampleg{direction}, \exampleg{left} $\Rightarrow$ \exampleg{gauche}, \exampleg{right} $\Rightarrow$ \exampleg{droite}, \exampleg{straight} $\Rightarrow$ \exampleg{tout}), \exampleg{back} $\Rightarrow$ \exampleg{arriere}, \exampleg{up} $\Rightarrow$ \exampleg{haut}, \exampleg{down} $\Rightarrow$ \exampleg{bas}, \exampleg{in} $\Rightarrow$ \exampleg{dans}, \exampleg{out} $\Rightarrow$ \exampleg{dehors}, \exampleg{on} $\Rightarrow$ \exampleg{sur}, \exampleg{under} $\Rightarrow$ \exampleg{sous}, \exampleg{behind} $\Rightarrow$ \exampleg{derriere}, \exampleg{next} $\Rightarrow$ \exampleg{prochain}, \exampleg{near} $\Rightarrow$ \exampleg{pres}, \exampleg{far} $\Rightarrow$ \exampleg{loin}, \exampleg{between} $\Rightarrow$ \exampleg{entre}, \exampleg{each} $\Rightarrow$ \exampleg{chacun}, \exampleg{all} $\Rightarrow$ \exampleg{tous}, \exampleg{some} $\Rightarrow$ \exampleg{quelques}, \exampleg{none} $\Rightarrow$ \exampleg{aucun}, \exampleg{every} $\Rightarrow$ \exampleg{chaque}, \exampleg{anyway} $\Rightarrow$ \exampleg{de toute facon} \exampleg{example} $\Rightarrow$ \exampleg{exemple}, \exampleg{reason} $\Rightarrow$ \exampleg{raison}, \exampleg{mistake} $\Rightarrow$ \exampleg{erreur}, \exampleg{gift} $\Rightarrow$ \exampleg{cadeau}, \exampleg{party} $\Rightarrow$ \exampleg{fete}, \exampleg{plan} $\Rightarrow$ \exampleg{plan}, \exampleg{goal} $\Rightarrow$ \exampleg{objectif}, \exampleg{success} $\Rightarrow$ \exampleg{succes}.   \\
         \bottomrule
    \end{tabular}
    \caption{200 Translation Pairs (English $\Rightarrow$ French) Used for ICL Capability Evaluation.}
    \label{tab:translation}
\end{table*}

\newpage
\twocolumn

\paragraph{Spelling Correction} We obtain the spelling correction data by prompting ChatGPT. Specifically, we use \exampleb{ChatGPT-o1}\footnote{\url{https://openai.com/index/openai-o1-system-card/}} with the prompt shown in Table~\ref{tab:chatgpt_prompts}(a). Table~\ref{tab:spell_corr} lists the generated spelling correction pairs. We then randomly sample 1,000 instances under three different settings (1-shot, 2-shot, and 5-shot) from these 200 pairs.

\paragraph{Translation} Similarly, for translation, we also sampled English to French word pairs by prompting \exampleb{ChatGPT-o1}. The prompt used is shown in Table~\ref{tab:chatgpt_prompts}(b). Table~\ref{tab:translation} lists the generated translation pairs. We then randomly sample 1,000 instances under three different settings (1-shot, 2-shot, and 5-shot) from these 200 pairs.

\section{General Capabilities Affected by Removal of Heads}
\label{app:more_effect}

We run the first 10 tasks from the MMLU benchmark using the \exampleb{Llama-3.1-8B} model with the deepeval\footnote{\url{https://www.deepeval.com/}} package under the default setting (no chain-of-thought). The entrainment heads are the same ones identified using the country--capital city relation, consistent with the rest of the evaluation in Section~\ref{sub:case_study}. Results are depicted in Table~\ref{tab:mmlu}.

\begin{table}
    \centering\small
    \begin{tabular}{lcc}
    \toprule
        \multirow{2}{*}{MMLU Task} & Before & After \\
         & Removing & Removing \\
    \midrule
        high school european history & 70.9 & 71.5 \\
        business ethics & 55.0 & 51.0 \\
        clinical knowledge & 58.4 & 58.8 \\
        medical genetics & 62.0 & 63.0 \\
        high school us history & 65.6 & 68.1 \\
        high school physics & 35.7 & 35.7 \\
        high school world history & 68.7 & 67.0 \\
        virology & 51.2 & 46.3 \\
        high school microeconomics & 58.4 & 57.1 \\
        econometrics & 36.8 & 36.8 \\
    \midrule
        Average & 56.3 & 55.5 \\
    \bottomrule
    \end{tabular}
    \caption{Effect of removing the entrainment heads on the MMLU tasks.}
    \label{tab:mmlu}
\end{table}

Overall, we see the same trend as in our evaluation on other LRE relations and the ICL tasks: a small decrease in overall performance (56.3\% to 55.5\%), but greater variance in the individual breakdown. There is no significant change in the model's performance, and the LMs' general ability remains in the same ballpark as the original model after removing the entrainment heads.
These results further support our hypothesis that this circuit of entrainment heads collectively corresponds more to contextual entrainment in specific settings, and less to the model's overall performance.

\end{document}